\documentclass[10pt,journal,compsoc]{IEEEtran}
\usepackage{multirow}
\usepackage{amsfonts}
\usepackage{amssymb}
\usepackage{amsmath}
\usepackage{subfigure}
\usepackage{graphicx}
\usepackage{mathrsfs}
\usepackage{algorithm}
\usepackage{algorithmic}

\def\iid{\stackrel{iid}{\sim}}
\def\DP{\mbox{DP}}
\def\iid{\stackrel{iid}{\sim}}

\def\1{\boldsymbol{1}}
\def\ii{\textbf{\textit{i}}}

\def\B{\mathcal{B}}
\def\E{\mathbb{E}}
\def\T{\mathcal{T}}
\def\I{\mathbb{I}}
\def\LB{\mathcal{L}}

\newcommand{\Dir}{\mbox{Dirichlet}}
\newcommand{\Beta}{\mbox{Beta}}

\begin{document}
\title{Nested Hierarchical Dirichlet Processes}

\author{John Paisley, Chong Wang, David M. Blei and Michael I. Jordan,~\IEEEmembership{Fellow,~IEEE}
\thanks{$\bullet$ John Paisley is with the Department of Electrical Engineering, Columbia University, New York, NY.

$\bullet$ Chong Wang is with Voleon Capital Management, Berkeley, CA.

$\bullet$ David M. Blei is with the Department of Computer Science, Princeton University, Princeton, NJ

$\bullet$ Michael I. Jordan is with the Departments of EECS and Statistics, UC Berkeley, Berkeley, CA}
}

\markboth{Journal of Pattern Analysis and Machine Intelligence,~Vol.~X, No.~X, XXXX}%
{Shell \MakeLowercase{\textit{et al.}}: Bare Demo of IEEEtran.cls for Computer Society Journals}

\IEEEtitleabstractindextext{
\begin{abstract}
We develop a nested hierarchical Dirichlet process (nHDP) for
hierarchical topic modeling. The nHDP generalizes the nested
Chinese restaurant process (nCRP) to allow each word to follow
its own path to a topic node according to a per-document distribution over the paths on a shared tree. This alleviates the rigid,
single-path formulation assumed by the nCRP, allowing documents
to easily express complex thematic borrowings. We derive a
stochastic variational inference algorithm for the model, which
enables efficient inference for massive collections of text
documents. We demonstrate our algorithm on 1.8 million documents
from \emph{The New York Times} and 2.7 million documents from
\emph{Wikipedia}.
\end{abstract}

}

\maketitle
\IEEEdisplaynontitleabstractindextext

\section{Introduction}
Organizing things hierarchically is a natural aspect of human
activity. Walking into a large department store, one might first
find the men's section, followed by men's casual, and then see
the t-shirts hanging along the wall. Or being hungry, one might
choose to eat Italian food, decide whether to spring for the
better, more authentic version or go to one of the cheaper chain
options, and then end up at the Olive Garden. Similarly with data
analysis, a hierarchical tree-structured representation of data
can provide an illuminating means for understanding and reasoning
about the information it contains.

In this paper, we focus on developing {\it hierarchical topic
models} to construct tree-structured representations for text
data.  Hierarchical topic models use a structured prior on the
topics underlying a corpus of documents, with the aim of bringing
more order to an unstructured set of thematic concepts
\cite{Blei:2003a}\cite{Wang:2009}\cite{Kim:2012}. They do this by
learning a tree structure for the underlying topics, with the
inferential goal being that topics closer to the root are more
general, and gradually become more specific in thematic content
when following a path down the tree.

\begin{figure*}[t]\centering
 \includegraphics[width=.8\textwidth]{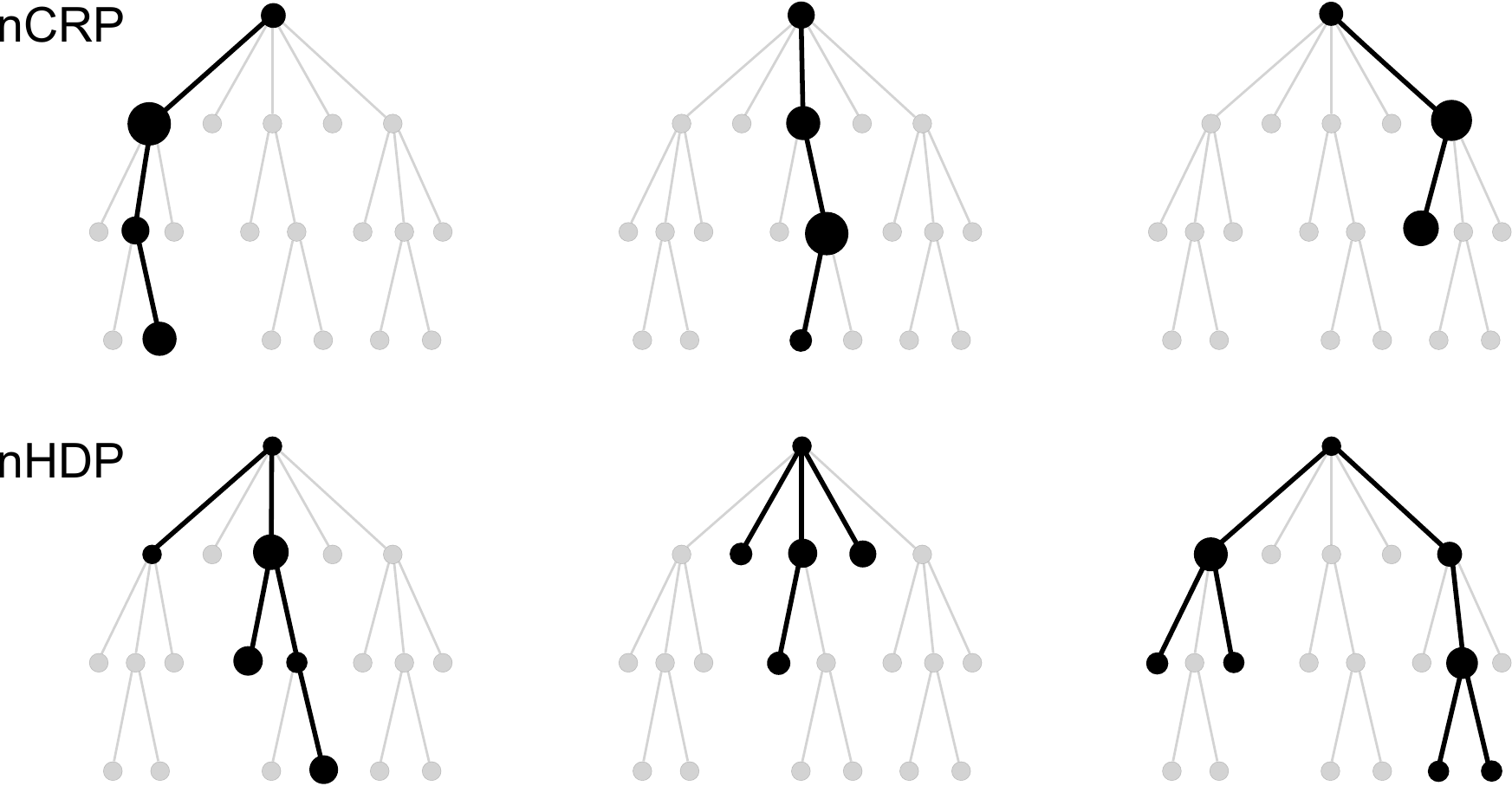}
\caption{An example of path structures for the nested Chinese
  restaurant process (nCRP) and the nested hierarchical Dirichlet
  process (nHDP) for hierarchical topic modeling. With the nCRP,
  the topics for a document are restricted to lying along a
  single path. With the nHDP, each document has
  access to the entire tree, but a document-specific distribution
  on paths will place high probability on a particular subtree. In both models a word follows a path to its topic. This path is deterministic in the case of the nCRP, and drawn from a highly probable document-specific subset of paths in the case of the nHDP.
  \label{fig.toy_trees}}
\end{figure*}

Our work builds on the nested Chinese restaurant process (nCRP)
\cite{Blei:2010}. The nCRP is a Bayesian nonparametric prior for
hierarchical topic models, but is limited in that it assumes each document selects topics from one path in the
tree. We illustrate this limitation in Figure
\ref{fig.toy_trees}. This assumption has practical drawbacks; for trees truncated to a small number of levels this does not allow for many topics per document, and for trees of many levels there are too many nodes to infer.  

The nCRP also has drawbacks from a modeling standpoint. As a simple example, consider an article on ESPN.com about an
injured player, compared with an article in a sports medicine
journal about a specific type of athletic injury. Both documents will contain words about medicine and
words about sports. These areas are different enough, however, that one cannot be considered to be a subset of the other. Yet the single-path structure of the nCRP will require this to be the case in order to model the relevant words in the documents, or it will learn a new ``sports/medicine'' topic rather than a mixture of separate sports and medicine topics. Continuing this analogy, other documents may only be about sports or medicine. As a result, medical terms in the nCRP will need to appear in multiple places within the tree: in its own subtree separate from sports, and also affiliated with sports, perhaps as a child of the general sports topic (in the case of the ESPN article). A similar fractionation of sports-related terms results from the sports medicine article, where the medical terms dominate and sports can be considered a topic underneath the main medicine topic. The result is a tree where topics appear in multiple places, and so the full statistical power
within the corpus is not being used to model each topic; the tree will not be as compact as it could be.

Though the nCRP is a Bayesian nonparametric prior, it performs
nonparametric clustering of \emph{document-specific} paths, which
reduces the number of topics available to a document by
restricting them to lie on a single path, leading to drawbacks as illustrated above. Our goal is to develop a
related Bayesian nonparametric prior that performs
\emph{word-specific} path clustering. We illustrate this
objective in Figure \ref{fig.toy_trees}. In this case, each word
has access to the entire tree, but with document-specific
distributions on the paths within the tree. To this end, we make
use of the hierarchical Dirichlet process \cite{Teh:2006},
developing a novel prior that we refer to as the \emph{nested
hierarchical Dirichlet process} (nHDP). The HDP can be viewed as
a nonparametric elaboration of the classical topic model, latent
Dirichlet allocation (LDA)~\cite{Blei:2003b}, providing a
mechanism whereby a global Dirichlet process defines a base
distribution for a collection of local Dirichlet
processes, one for each document. With the nHDP, we extend this idea by letting a global
nCRP become a base distribution for a collection of local
nCRPs, one for each document. As illustrated in Figure \ref{fig.toy_trees}, the nested HDP provides the
opportunity for cross-thematic borrowing while keeping general topic areas in separate subtrees, which is not possible
with the nCRP. 

Hierarchical topic models have thus far been applied to corpora
of small size. A significant issue, not just with topic models
but with Bayesian models in general, is to scale up inference to
massive data sets \cite{Jordan:2011}. Recent developments in
stochastic variational inference methods have shown promising
results for LDA and the HDP topic model
\cite{Hoffman:2012}\cite{Hoffman:2010}\cite{Wang:2011}. We
continue this development for hierarchical topic modeling with the
nested HDP. Using stochastic variational inference, we
demonstrate an ability to efficiently handle very large corpora.
This is a major benefit to complex models such as tree-structured
topic models, which require significant amounts of data to
support their large size.

We organize the paper as follows: In Section \ref{sec.review} we
review the Bayesian nonparametric priors that we incorporate in
our model---the Dirichlet process, nested Chinese restaurant
process and hierarchical Dirichlet process. In Section
\ref{sec.nHDP} we present our proposed nested HDP model for
hierarchical topic modeling. In Section \ref{sec.inference} we
review stochastic variational inference and present an inference
algorithm for nHDPs that scales well to massive data sets. We
present empirical results in Section \ref{sec.experiments}. We
first compare the nHDP with the nCRP on three relatively small
data sets. We then evaluate our stochastic algorithm on 1.8
million documents from \emph{The New York Times} and 2.7 million
documents from \emph{Wikipedia}, comparing performance with
stochastic LDA and HDP.

\section{Background: Bayesian nonparametric priors for topic models}\label{sec.review}
The nested hierarchical Dirichlet process (nHDP) builds on a collection of existing Bayesian nonparametric priors. In this section, we review these priors: the Dirichlet process, nested Chinese restaurant process and hierarchical Dirichlet process. We also review constructive representations for these processes that we will use for posterior inference of the nHDP topic model.

\subsection{Dirichlet processes}
The Dirichlet process (DP) \cite{Ferguson:1973} is the foundation for a large collection of Bayesian nonparametric models that rely on mixtures to represent distributions on data. Mixture models work by partitioning a data set according to statistical traits shared by members of the same cell. Dirichlet process priors are effective in learning a suitable number of traits for representing the data, in addition to the parameters of the mixture. The basic form of a Dirichlet process mixture model is
\begin{equation}\label{eqn.mixture}
 W_n | \varphi_n \sim F_W(\varphi_n),\quad \varphi_n | G \iid G, \quad G = \sum_{i=1}^{\infty} p_i \delta_{\theta_i}.
\end{equation}
With this representation, data $W_1,\dots,W_N$ are distributed according to a family of distributions $F_W$ with respective parameters $\varphi_1,\dots,\varphi_N$. These parameters are drawn from the distribution $G$, which is discrete and potentially infinite, as the DP allows it to be. This discreteness induces a partition of the data $W$ according to the sharing of the atoms $\{\theta_i\}$ among the parameters $\{\varphi_n\}$ that are selected.

The Dirichlet process is a stochastic process for generating $G$. To briefly review, let $(\Theta,\B)$ be a measurable space, $G_0$ a probability measure on it and $\alpha > 0$. Ferguson \cite{Ferguson:1973} proved the existence of a stochastic process $G$ where, for all measurable partitions $\{B_1,\dots,B_k\}$ of $\Theta$, with $B_i \in \mathcal{B}$,
$$(G(B_1),\dots,G(B_k)) \sim \Dir(\alpha G_0(B_1),\dots,\alpha G_0(B_k)),$$
abbreviated as $G\sim\DP(\alpha G_0)$. It has been shown that $G$ is discrete (with probability one) even when $G_0$ is non-atomic \cite{Blackwell:1973}\cite{Sethuraman:1994}. Thus the DP prior is a good candidate for $G$ in Eq.\ (\ref{eqn.mixture}) since it generates discrete distributions on continuous parameter spaces.  For most applications $G_0$ is diffuse, and so representations of $G$ at the granularity of the atoms are necessary for inference; we next review two of these approaches to working with this infinite-dimensional distribution.

\subsubsection{Chinese restaurant processes} 
The Chinese restaurant process (CRP) avoids directly working with $G$ by integrating it out \cite{Blackwell:1973}\cite{Aldous:1985}. In doing so, the values of $\varphi_1,\dots,\varphi_N$ become dependent, with the value of $\varphi_{n+1}$ given $\varphi_1,\dots,\varphi_n$ distributed as
\begin{equation}\label{eqn.CRP}
 \varphi_{n+1}|\varphi_1,\dots,\varphi_n \sim \frac{\alpha}{\alpha+n}G_0 + \sum_{i=1}^n \frac{1}{\alpha + n} \delta_{\varphi_i}.
\end{equation}
That is, $\varphi_{n+1}$ takes the value of one of the previously observed $\varphi_i$ with probability $\frac{n}{\alpha+n}$, and a value drawn from $G_0$ with probability $\frac{\alpha}{\alpha+n}$, which will be unique when $G_0$ is continuous. This displays the clustering property of the CRP and also gives insight into the impact of $\alpha$, since it is evident that the number of unique $\varphi_i$ grows like $\alpha\ln n$. In the limit $n\rightarrow\infty$, the distribution in Eq.\ (\ref{eqn.CRP}) converges to a random measure distributed according to a Dirichlet process \cite{Blackwell:1973}. The CRP is so-called because of an analogy to a Chinese restaurant, where a new customer (datum) sits at a table (selects a parameter) with probability proportional to the number of previous customers at that table, or selects a new table with probability proportional to $\alpha$.

\subsubsection{A stick-breaking construction} 
Where the Chinese restaurant process works with $G\sim\DP(\alpha G_0)$ implicitly through $\varphi$, a stick-breaking construction allows one to directly construct $G$ before drawing any $\varphi_n$. Sethuraman \cite{Sethuraman:1994} showed that if $G$ is constructed as follows:
$$G = \sum_{i=1}^{\infty}V_i\prod_{j=1}^{i-1}(1-V_j)\delta_{\theta_i},$$
\begin{equation}\label{eqn.stick}
V_i \iid \Beta(1,\alpha),\quad \theta_i \iid G_0,
\end{equation}
then $G\sim\DP(\alpha G_0)$. The variable $V_i$ can be interpreted as the proportion broken from the remainder of a unit length stick, $\prod_{j<i}(1-V_j)$. As the index $i$ increases, more random variables in $[0,1]$ are multiplied, and thus the weights decrease to zero exponentially. The expectation $\E[V_i\prod_{j<i}(1-V_j)] = \frac{\alpha^{i-1}}{(1+\alpha)^i}$ gives a sense of the impact of $\alpha$ on these weights. This explicit construction of $G$ maintains the independence among $\varphi_1,\dots,\varphi_N$ as written in Eq.\ (\ref{eqn.mixture}), which is a significant advantage of this representation for mean-field variational inference that is not present in the CRP.

\subsection{Nested Chinese restaurant processes}
Nested Chinese restaurant processes (nCRP) are a tree-structured extension of the CRP that are useful for hierarchical topic modeling \cite{Blei:2010}. They extend the CRP analogy to a nesting of restaurants in the following way: After selecting a table (parameter) according to a CRP, the customer departs for another restaurant uniquely indicated by that table. Upon arrival, the customer acts according to the CRP for the new restaurant, and again departs for a restaurant only accessible through the table selected. This occurs for a potentially infinite sequence of restaurants, which generates a sequence of parameters for the customer according to the selected tables.

A natural interpretation of the nCRP is as a tree where each parent has an infinite number of children. Starting from the root node, a path is traversed down the tree. Given the current node, a child node is selected with probability proportional to the previous number of times it was selected among its siblings, or a new child is selected with probability proportional to $\alpha$. As with the CRP, the underlying mixing measure of the nCRP also has a constructive representation useful for variational inference, which we will use in our nHDP construction.

\subsubsection{Constructing the nCRP}\label{sec.nCRPstick}
The nesting of Dirichlet processes that leads to the nCRP gives rise to a stick-breaking construction \cite{Wang:2009}. We develop the notation for this construction here and use it later in our construction of the nested HDP. Let $\ii_l = (i_1,\dots,i_l)$ be a path to a node at level $l$ of the tree.\footnote{That is, from the root node first select the child with index $i_1$; from node $\ii_1 = (i_1)$, select the child with index $i_2$; from node $\ii_2 = (i_1,i_2)$ select the child with index $i_3$, and so on to level $l$ with each $i_k \in \mathbb{N}$. We ignore the root $\ii_0$, which is shared by all paths.} According to the stick-breaking version of the nCRP, the children of node $\ii_l$ are countably infinite, with the probability of transitioning to child $j$ equal to the $j$th break of a stick-breaking construction. 
Each child corresponds to a parameter drawn independently from $G_0$. Letting the index of the parameter identify the index of the child, this results in the following DP for the children of node $\ii_l$,
$$G_{\ii_l} = \sum_{j=1}^{\infty} V_{\ii_l,j}\prod_{m=1}^{j-1}(1-V_{\ii_l,m})\delta_{\theta_{(\ii_l,j)}},$$
\begin{equation}\label{eqn.nCRP_stick_top}
V_{\ii_l,j} \iid \Beta(1,\alpha),\quad \theta_{(\ii_l,j)} \iid G_0.
\end{equation}
If the next node is child $j$, then the nCRP transitions to DP $G_{\ii_{l+1}}$, where $\ii_{l+1}$ has index $j$ appended to $\ii_l$, that is $\ii_{l+1} = (\ii_l,j)$. A path down the tree givens a sequence of parameters $\boldsymbol{\varphi} = (\varphi_1,\varphi_2,\dots)$, where the parameter $\varphi_l$ correspond to an atom $\theta_{\ii_l}$ at level $l$. Hierarchical topic models use these sequences of parameters to give the topics for generating documents. Other nested DPs have been considered as well, such as a two-leveled nDP where all parameters are selected from the leaves \cite{Rodriguez:2008}.

\subsubsection{Nested CRP topic models}\label{sec.nCRPdoc}
Hierarchical topic models based on the nested CRP use a globally shared tree to generate a corpus of documents. Starting with the construction of nested Dirichlet processes as described above, each document selects a path down the tree according to a Markov process, which produces a sequence of topics $\boldsymbol{\varphi}_d = (\varphi_{d,1},\varphi_{d,2},\dots)$ used to generate the $d$th document. As with other topic models, each word in a document, $W_{d,n}$, is represented by an index in the set $\{1,\dots,\mathcal{V}\}$ and the topics $\theta_{\ii_l}$ appearing in $\boldsymbol{\varphi}_d$ are $\mathcal{V}$-dimensional probability vectors with Dirichlet prior $G_0 = \mbox{Dirichlet}(\lambda_0\boldsymbol{1}_{\mathcal{V}})$.

For each document $d$, an additional stick-breaking process provides a distribution on the topics in $\boldsymbol{\varphi}_d$,
\begin{eqnarray}\label{eqn.level_dist}
 G^{(d)} &=& \sum_{j=1}^{\infty} U_{d,j}\prod_{m=1}^{j-1}(1-U_{d,m})\delta_{\varphi_{d,j}},\nonumber\\
U_{d,j} &\iid& \Beta(\gamma_1,\gamma_2).
\end{eqnarray}
Since this is not a DP, $U_{d,j}$ has two free parameters, $\gamma_1$ and $\gamma_2$.  Following the standard method, words for document $d$ are generated by first drawing a topic i.i.d.\ from $G^{(d)}$, and then drawing the word index from the discrete distribution with the selected topic.

\subsubsection{Issues with the nCRP} As discussed in the introduction, a significant drawback of the nCRP for topic modeling is that each document follows one path down the tree. Therefore, all thematic content of a document must be contained within that single sequence of topics. Since the nCRP is meant to characterize the thematic content of a corpus in increasing levels of specificity, this creates a combinatorial problem, where similar topics will appear in many parts of the tree to account for the possibility that they appear as a topic of the document (e.g., the sport/medicine example given in the introduction). In practice, nCRP trees are typically truncated at three levels \cite{Wang:2009}\cite{Blei:2010}, since learning deeper levels becomes difficult due to the exponential increase in nodes.\footnote{This includes a root node topic, which is shared by all documents and is intended to collect stop words.} In this situation each document has three topics for modeling its entire thematic content, 
which is likely insufficient, and so a blending of multiple topics is bound to occur during inference.

The nCRP is a Bayesian nonparametric (BNP) prior, but it performs nonparametric clustering of the paths selected at the document level, rather than at the word level. Though the same distribution on a tree is shared by a corpus, each document can differentiate itself only by the path it choses, as well as the distribution on topics in that path. The key issue with the nCRP is the restrictiveness of this single path allowed to a document. However, if instead each word were allowed to follow its own path according to an nCRP, the distribution on paths would be the same for all documents, which is clearly not desired. Our goal is to develop a hierarchical topic model that does not prohibit a document from using topics in different parts of the tree. Our solution to this problem is to employ the hierarchical Dirichlet process (HDP).

\subsection{Hierarchical Dirichlet processes}\label{sec.HDP}
The HDP is a multi-level version of the Dirichlet process \cite{Teh:2006}. It makes use of the idea that the base distribution on the continuous space $\Theta$ can be discrete, which is useful because a discrete distribution allows for multiple draws from the DP prior to place probability mass on the same subset of atoms. Hence different groups of data can share the same atoms, but have different probability distributions on them. A discrete base is needed, but the atoms are unknown in advance. The HDP models these atoms by drawing the base from a DP prior. This leads to the hierarchical process
\begin{equation}
 G_d | G \iid \DP(\beta G),\quad G \sim \DP(\alpha G_0),
\end{equation}
for groups $d = 1,\dots,D$. This prior has been used to great effect in topic modeling as a nonparametric extension of LDA \cite{Blei:2003b} and related LDA-based models \cite{Ren:2008}\cite{Airoldi:2008}\cite{Fox:2011}.

As with the DP, explicit representations of the HDP are necessary for inference. The representation we use relies on two levels of Sethuraman's stick breaking construction. For this construction, first sample $G$ as in Eq.\ (\ref{eqn.stick}), and then sample $G_d$ in the same way,
$$G_d = \sum_{i=1}^{\infty} V_i^d\prod_{j=1}^{i-1}(1-V_j^d)\delta_{\phi_i},$$
\begin{equation}
V_i^d \iid \Beta(1,\beta),\quad \phi_i \iid G.
\end{equation}
This form is identical to Eq.\ (\ref{eqn.stick}), with the key difference that $G$ is discrete, and so atoms $\phi_i$ will repeat. An advantage of this representation is that all random variables are i.i.d., which aids variational inference strategies.

\section{Nested hierarchical Dirichlet processes for topic modeling}\label{sec.nHDP}
In building on the nCRP framework, our goal is to allow for each document to have access to the entire tree, while still learning document-specific distributions on topics that are thematically coherent. Ideally, each document will still exhibit a dominant path corresponding to its main themes, but with off-shoots allowing for other topics. Our two major changes to the nCRP formulation toward this end are that ($i$) each word follows its own path to a topic, and ($ii$) each document has its own distribution on paths in a shared tree. The BNP tools discussed above make this a straightforward task. 

In the proposed nested hierarchical Dirichlet process (nHDP), we split the process of generating a document's distribution on topics into two parts: first, generating a document's distribution on paths down the tree, and second, generating a word's distribution on terminating at a particular node within those paths.

\subsection{Constructing a distribution on paths}
With the nHDP, all documents share a global nCRP drawn according to the stick-breaking construction in Section \ref{sec.nCRPstick}. Denote this tree by $\T$. As discussed, $\T$ is simply an infinite collection of Dirichlet processes with a continuous base distribution $G_0$ and a transition rule between DPs. According to this rule, from a root Dirichlet process $G_{\ii_0}$, a path is followed by drawing $\varphi_{l+1} \sim G_{\ii_l}$ for $l = 0, 1, 2, \dots$, where $\ii_0$ is a constant root index, and $\ii_l = (i_1,\dots,i_l)$ indexes the DP associated with the topic $\varphi_l = \theta_{\ii_l}$. With the nested HDP, instead of following paths according to the global $\T$, we use each Dirichlet process in $\T$ as a base distribution for a local DP drawn independently for each document.

That is, for document $d$ we construct a tree $\T_d$ where, for each $G_{\ii_l} \in \T$, we draw a corresponding $G^{(d)}_{\ii_l} \in \T_d$ according to the Dirichlet process
\begin{equation}\label{eqn.nHDPlevel2}
G^{(d)}_{\ii_l} \sim \DP(\beta G_{\ii_l}).
\end{equation}
As discussed in Section \ref{sec.HDP}, $G^{(d)}_{\ii_l}$ will have the same atoms as $G_{\ii_l}$, but with different probability weights on them. Therefore, the tree $\T_d$ will have the same nodes as $\T$, but the probability of a path in $\T_d$ will vary with $d$, giving each document its own distribution on the tree.

We represent this document-specific DP with a stick-breaking construction as in Section \ref{sec.HDP},
$$G^{(d)}_{\ii_l} = \sum_{j=1}^{\infty}V^{(d)}_{\ii_l,j}\prod_{m=1}^{j-1}(1-V^{(d)}_{\ii_l,m})\delta_{\phi^{(d)}_{\ii_l,j}},$$
\begin{equation}\label{eqn.nHDP_stick_second}
V^{(d)}_{\ii_l,j} \iid \Beta(1,\beta),\quad \phi^{(d)}_{\ii_l,j} \iid G_{\ii_l}.
\end{equation}
This representation retains full independence among random variables, and will lead to a simpler stochastic variational inference algorithm. We note that the atoms from the global DP are randomly permuted and copied with this construction; $\phi^{(d)}_{\ii_l,j}$ does not correspond to the node with parameter $\theta_{(\ii_l,j)}$. To find the probability mass that $G^{(d)}_{\ii_l}$ places on $\theta_{(\ii_l,j)}$, one can calculate $$G^{(d)}_{\ii_l}(\{\theta_{(\ii_l,j)}\}) = \textstyle\sum_m G^{(d)}_{\ii_l}(\{\phi^{(d)}_{\ii_l,m}\})\I(\phi^{(d)}_{\ii_l,m}=\theta_{(\ii_l,j)}).$$

Using this nesting of HDPs to construct $\T_d$, each document has a tree with transition probabilities defined over the same subset of nodes since $\T$ is discrete, but with values for these probabilities that are document specific. To see how this allows each word to follow its own path while still producing a thematically coherent document, consider each $G^{(d)}_{\ii_l}$ when $\beta$ is small. In this case, most of the probability will be placed on one atom selected from $G_{\ii_l}$ since the first proportion $V^{(d)}_{\ii_l,1}$ will be large with high probability. This will leave little probability remaining for other atoms, a feature shared by all DPs in $\T_d$. Starting from the root node of $\T_d$, each word in the document will have high probability of transitioning to the same node when moving down the tree, with some small probability of diverging into a different topic. In the limit $\beta \rightarrow 0$, each $G^{(d)}_{\ii_l}$ will be a delta function on a $\phi^{(d)}_{\ii_l,j} \sim G_{\ii_l}$, 
and the 
same path will be selected by each word with probability one, thus recovering the nCRP.

\begin{algorithm}[t]
   \caption{Generating documents with the nHDP}
   \label{alg.nHDP}
\raggedright\vspace{4pt}
\begin{enumerate}
\item Generate a global tree $\T$ by constructing an nCRP as in Section \ref{sec.nCRPstick}.
\item Generate document tree $\T_d$ and switching probabilities $\boldsymbol{U}^{(d)}$. For document $d$,\\
  $\quad$ a) For each DP in $\T$, draw a DP with this as a\\$\quad~~~\,$ base distribution (Equation \ref{eqn.nHDPlevel2}).\\
  $\quad$ b) For each node in $\T_d$, draw a beta random\\$\quad~~~\,$  variable (Equation \ref{eqn.nHDPbeta}).\\
\item Generate a document. For word $n$ in document $d$,\\
  $\quad$ a) Sample atom $\varphi_{d,n}$ (Equation \ref{eqn.samp_node}).\\
  $\quad$ b) Sample word $W_{d,n}$ from topic $\varphi_{d,n}$.
\end{enumerate}
\end{algorithm}

\subsection{Generating a document}
With the tree $\T_d$ for document $d$ we have a method for selecting word-specific paths that are thematically coherent, meaning they tend to reuse the same path while allowing for off-shoots. We next discuss how to generate a document with this tree. As discussed in Section \ref{sec.nCRPdoc}, with the nCRP the atoms selected for a document by its path through $\T$ have a unique stick-breaking distribution that determines which level any particular word comes from.  We generalize this idea to the tree $\T_d$ with an overlapping stick-breaking construction as follows.

For each node $\ii_l$, we draw a document-specific beta random variable that acts as a stochastic switch. Given a pointer that is currently at node $\ii_l$, the beta random variable determines the probability that we draw from the topic at that node or continue further down the tree. That is, given that the path for word $W_{d,n}$ is at node $\ii_l$, stop with probability $U_{d,\ii_l}$, where
\begin{equation}\label{eqn.nHDPbeta}
 U_{d,\ii_l} \sim \Beta(\gamma_1,\gamma_2).
\end{equation}
If we don't select topic $\theta_{\ii_l}$, then continue by selecting node $\ii_{l+1}$ according to $G^{(d)}_{\ii_l}$. We observe the stick-breaking construction implied by this construction; for word $n$ in document $d$, the probability that its topic $\varphi_{d,n} = \theta_{\ii_l}$ is
\begin{equation}\label{eqn.samp_node}
 \mbox{Pr}(\varphi_{d,n} = \theta_{\ii_l}|\mathcal{T}_d,\boldsymbol{U}_d) = \hspace{1.5in}
\end{equation}
$$\left[\prod_{m=0}^{l-1} G^{(d)}_{\ii_m}\left(\lbrace\theta_{\ii_{m+1}}\rbrace\right)\right]\left[U_{d,\ii_l}\prod_{m=1}^{l-1}(1-U_{d,\ii_m})\right].$$
Here it is implied that $\ii_m$ equals the first $m$ values in $\ii_l$ for $m \leq l$. The leftmost term in this expression is the probability of path $\ii_l$, the right term is the probability that the word does not select the first $l-1$ topics, but does select the $l$th. Since all random variables are independent, a simple product form results that will significantly aid the development of a posterior inference algorithm. The overlapping nature of this stick-breaking construction on the levels of a sequence is evident from the fact that the random variables $U$ are shared for the first $l$ values by all paths along the subtree starting at node $\ii_l$. A similar tree-structured prior distribution was presented by Adams, et al. \cite{Adams:2010} in which all groups shared the same distribution on a tree and entire objects (e.g., images or documents) were clustered within a single node. We summarize our model for generating documents with the nHDP in Algorithm \ref{alg.nHDP}.

\section{Stochastic variational inference for the Nested HDP}\label{sec.inference}
Many text corpora can be viewed as ``Big Data''---they are large data sets for which standard inference algorithms can be prohibitively slow. For example, \emph{Wikipedia} currently indexes several million entries and \emph{The New York Times} has published almost two million articles in the last 20 years. With so much data, fast inference algorithms are essential. Stochastic variational inference is a development in this direction for hierarchical Bayesian models in which ideas from stochastic optimization are applied to approximate Bayesian inference using mean-field variational Bayes (VB) \cite{Sato:2001}\cite{Hoffman:2012}. Stochastic inference algorithms have provided a significant speed-up in inference for probabilistic topic models~\cite{Hoffman:2010}\cite{Wang:2011}\cite{Paisley:2012}.  In this section, after reviewing the ideas behind stochastic variational inference, we present a stochastic variational inference algorithm for the nHDP topic model.

\subsection{Stochastic variational inference}
Stochastic variational inference exploits the difference between \emph{local} variables, or those associated with a single unit of data, and \emph{global} variables, which are shared over an entire data set. In brief, stochastic VB works by splitting a large data set into smaller groups, processing the local variables of one group, updating the global variables, and then moving to another group. This is in contrast to batch inference, which processes all local variables at once before updating the global variables. In the context of probabilistic topic models, the unit of data is a document, and the global variables include the topics (among other possible variables), while the local variables relate to the distribution on these topics for each document. We next briefly review the relevant ideas from variational inference and its stochastic variant. 

\subsubsection{The batch set-up} Mean-field variational inference is a method for approximate posterior inference in Bayesian models \cite{Jordan:1999}. It approximates the full posterior of a set of model parameters $P(\Phi|W)$ with a factorized distribution $Q(\Phi|\Psi) = \prod_i q_i(\phi_i|\psi_i)$. It does this by searching the space of variational approximations for one that is close to the posterior according to their Kullback-Leibler divergence. Algorithmically, this is done by maximizing a variational objective function $\LB$ with respect to the variational parameters $\Psi$ of $Q$, where
\begin{equation}\label{eqn.LB_generic}
 \LB(W,\Psi) = \E_Q[\ln P(W,\Phi)] - \E_Q[\ln Q].
\end{equation}

We are interested in conjugate exponential models, where the prior and likelihood of all nodes of the model fall within the conjugate exponential family. In this case, variational inference has a simple optimization procedure \cite{Winn:2005}, which we illustrate with the following example---this generic example gives the general form exploited by the stochastic variational inference algorithm that we apply to the nHDP.

Consider $D$ independent samples from an exponential family distribution $P(W|\eta)$, where $\eta$ is the natural parameter vector. The likelihood under this model has the generic form
\begin{equation}\nonumber
P(W_{1:D}|\eta) = \left[\prod_{d=1}^D h(W_d)\right]\exp\left\{ \eta^T \sum_{d=1}^D t(W_d) - D A(\eta)\right\}.
\end{equation}
The sum of vectors $t(W_d)$ forms the sufficient statistics of the likelihood. The conjugate prior on $\eta$ has a similar form
\begin{equation}\nonumber
P(\eta|\chi,\nu) = f(\chi,\nu)\exp\left\{ \eta^T\chi - \nu A(\eta)\right\}.
\end{equation}
Conjugacy between these two distributions motivates selecting a $q$ distribution in this same family to approximate the posterior of $\eta$,
\begin{equation}\nonumber
q(\eta|\chi',\nu') = f(\chi',\nu')\exp\left\{ \eta^T\chi' - \nu' A(\eta)\right\}.
\end{equation}
The variational parameters $\chi'$ and $\nu'$ are free and are modified to maximize the lower bound in Eq.\ (\ref{eqn.LB_generic}).\footnote{A closed form expression for the lower bound is readily derived for this example.} Inference proceeds by taking the gradient of $\LB$ with respect to the variational parameters of a particular $q$, in this case the vector $\psi:=[\chi'^T,\nu']^T$, and setting to zero to find their updated values. For the conjugate exponential example we are considering, this gradient is
\begin{equation}\label{eqn.LBgrad}
\nabla_{\psi} \LB(W,\Psi) = -\left[\begin{matrix}\frac{\partial^2\ln f}{\partial\chi'\partial\chi'^T} & \frac{\partial^2\ln f}{\partial\chi'\partial\nu'}\vspace{3mm} \\ \frac{\partial^2\ln f}{\partial\nu'\partial\chi'^T} & \frac{\partial^2\ln f}{\partial\nu'^2} \end{matrix}\right]\left[\begin{matrix}\chi + \displaystyle\textstyle\sum_d t_d - \chi'\vspace{2mm}\\ \nu+D - \nu'\end{matrix}\right].
\end{equation}
Setting this to zero, one can immediately read off the variational parameter updates from the rightmost vector. In this case $\chi' = \chi + \sum_{d=1}^D t(W_d)$ and $\nu' = \nu + D$, which are the sufficient statistics calculated from the data.

\subsubsection{A stochastic extension}\label{sec.stoch}
 Stochastic optimization of the variational lower bound modifies batch inference by forming a noisy gradient of $\LB$ at each iteration. The variational parameters for a random subset of the data are optimized first, followed by a step in the direction of the noisy gradient of the global variational parameters. Let $C_s \subset \{1,\dots,D\}$ index a subset of the data at step $s$. Also let $\phi_d$ be the hidden local variables associated with observation $W_d$ and let $\Phi_W$ be the global variables shared among all observations. The stochastic variational objective function $\LB_s$ is the noisy version of $\LB$ formed by selecting a subset of the data,
$$\LB_s(W_{C_s},\Psi) = \frac{D}{|C_s|} \sum_{d\in C_s}\E_Q[\ln P(W_d,\phi_d|\Phi_{W})] $$
\begin{equation}\label{eqn.stochL}
  \hspace{1in}+~ \E_Q[\ln P(\Phi_{W}) -\ln Q].
\end{equation}
Optimizing $\LB_s$ optimizes $\LB$ in expectation; since each subset $C_s$ is equally probable, with $p(C_s) = {D \choose |C_s|}^{-1}$, and since $d\in C_s$ for ${D-1 \choose |C_s|-1}$ of the ${D \choose |C_s|}$ possible subsets, it follows that $ \E_{p(C_s)}[\LB_s(W_{C_s},\Psi)] = \LB(W,\Psi).$

Stochastic variational inference proceeds by optimizing the objective in (\ref{eqn.stochL}) with respect to $\psi_d$ for $d \in C_s$, followed by an update to $\Psi_W$ that blends the new information with the old. The update of a global variational parameter $\psi$ at step $s$ is $\psi_s = \psi_{s-1} + \rho_s B\nabla_{\psi} \LB_s(W_{C_s},\Psi)$, where the matrix $B$ is a positive definite preconditioning matrix and $\rho_s$ is a step size satisfying $\sum_{s=1}^{\infty}\rho_s = \infty$ and $\sum_{s=1}^{\infty}\rho_s^2 < \infty$ to ensure convergence \cite{Sato:2001}.

The gradient $\nabla_{\psi} \LB_s(W_{C_s},\Psi)$ has a similar form as Eq.\ (\ref{eqn.LBgrad}), with the exception that the sum is taken over a subset of the data. Though the matrix in Eq.\ (\ref{eqn.LBgrad}) is often very complicated, it is superfluous to batch variational inference for conjugate exponential family models. In the stochastic optimization of Eq.\ (\ref{eqn.LB_generic}), however, this matrix cannot be ignored. The key for conjugate exponential models is in selecting the preconditioning matrix $B$. Since the gradient of $\LB_s$ has the same form as Eq.\ (\ref{eqn.LBgrad}), $B$ can be set to the inverse of the matrix in (\ref{eqn.LBgrad}) to allow for cancellation. An interesting observation is that this matrix is 
\begin{equation}
B = -\left(\frac{\partial^2 \ln q(\eta|\psi)}{\partial \psi\partial \psi^T}\right)^{-1},
\end{equation}
which is the inverse Fisher information of the variational distribution $q(\eta|\psi)$. Using this setting for $B$, the step direction is the natural gradient of the lower bound, and therefore  gives an efficient step direction in addition to simplifying the algorithm \cite{Amari:1998}. The resulting variational update is a weighted combination of the old sufficient statistics for $q$ with the new ones calculated over data indexed by $C_s$.

\begin{algorithm}[t]
   \caption{Variational inference for the nHDP}
   \label{alg.vbinf}
\raggedright\vspace{4pt}
\begin{enumerate}
\item Randomly subsample documents from the corpus.
\item For each document in the subsample,\\
  $\quad$ a) Select a subtree according to a greedy\\$\quad~~~\,$ process on the variational objective (Eq.\ \ref{eq.greedy_select}).\\
  $\quad$ b) Optimize $q$ distributions for subtree.\\$\quad~~~\,$ Iterate between word allocation (Eq.\ \ref{eqn.upC})\\$\quad~~~\,$ and topic distribution updates (Eqs.\ \ref{eqn.upUV}--\ref{eqn.upAB}).\\
\item Collect the sufficient statistics for the topics and base distribution and step in the direction of the natural gradient (Eqs.\ \ref{eqn.upTheta}--\ref{eqn.upTau}).
\item Return to Step 1.
\end{enumerate}
\end{algorithm}

\subsection{The inference algorithm}
We develop a stochastic variational inference algorithm for approximate posterior inference of the nHDP topic model. As discussed in our general review of stochastic inference, this entails optimizing the local variational parameters for a subset of documents, followed by a step along the natural gradient of the global variational parameters. We distinguish between local and global variables for the nHDP in Table \ref{tab.q_dist}. In Table \ref{tab.q_dist} we also give the variational $q$ distributions selected for each variable. In almost all cases we select this distribution to be in the same family as the prior. We point out two additional latent indicator variables for inference: $c_{d,n}$, which indicates the topic of word $W_{d,n}$, and $z_{\ii,j}^{(d)}$, which points to the atom in $G_{\ii}$ associated with the $j$th break in $G^{(d)}_{\ii}$ using the construction given in Eq.\ (\ref{eqn.nHDP_stick_second}).

Since we wish to consider large trees, and because there is slightly more overhead in calculating the distribution for each document than in models such as LDA and the HDP, the word allocation step is more time consuming for the nHDP. Additionally, we seek an efficient means for learning the indicators $z_{\ii,j}^{(d)}$. Since each document will use a small subset of topics, which translates to a small subtree of the entire tree, our goal is to pick out a subtree in advance for the document to work with. This will reduce the number of topics to do inference over for each document, speeding up the algorithm, and determine the delta-function indicators for $z_{\ii,j}^{(d)}$, which point to the ``activated'' nodes.

To this end, we introduce a third aspect to our inference algorithm in which we pick a small subtree for each document in advance. By this we mean that we only allow words in a document to be allocated to the subtree selected for that document and fix the probability that the indicator $c_{d,n}$ corresponds to topics outside this subtree to zero. As we will show, by selecting a subtree we are in effect learning a truncated stick-breaking construction of the tree for each document. If a node has two children in the subtree, then algorithmically we will have a two-node truncated construction for that DP of the {\it specific} document we are considering.

We select the subtree from $\T$ for each document using a greedy algorithm. This greedy algorithm is performed with respect to maximizing the variational objective function. Being an optimization method with one requirement (that we maximize a fixed objective), variational inference has considerable freedom in this regard. We discuss this greedy algorithm below, followed by the variational parameter updates for the local and global $q$ distributions. Algorithm 2 gives an outline.

\begin{table*}
\begin{center}\caption{A list of the local and global variables and their respective $q$ distributions for the nHDP topic model.}\label{tab.q_dist}
\begin{tabular}{rrll}
\hline
Global variables: & $\theta_{\ii}$&\hspace{-3mm}: topic probability vector for node $\ii$ & $q(\theta_{\ii}) = \Dir(\theta_{\ii}|\lambda_{\ii,1},\dots,\lambda_{\ii,\mathcal{V}})$\\
		  & $V_{\ii,j}$ &\hspace{-3mm}: stick proportion for the global DP for node $\ii$ & $q(V_{\ii,j}) = \Beta(V_{\ii,j}|\tau^{(1)}_{\ii,j},\tau^{(2)}_{\ii,j})$\\
Local variables:  & $V^{(d)}_{\ii,j}$ &\hspace{-3mm}: stick proportion for local DP for node $\ii$ & $q(V_{\ii,j}) = \Beta(V^{(d)}_{\ii,j}|u^{(d)}_{\ii,j},v^{(d)}_{\ii,j})$\\
		  & $z^{(d)}_{\ii,j}$ &\hspace{-3mm}: index pointer to atom in $G_{\ii}$ for $j$th break in $G^{(d)}_{\ii}$ & $q(z^{(d)}_{\ii,j}) = \delta_{z^._{.,.}}(k)$, $k = 1,2,\dots$\\
		  & $U_{d,\ii}$ &\hspace{-3mm}: beta distributed switch probability for node $\ii$ & $q(U_{d,\ii}) = \Beta(U_{d,\ii}|a_{d,\ii},b_{d,\ii})$\\
		  & $c_{d,n}$ &\hspace{-3mm}: topic indicator for word $n$ in document $d$ & $q(c_{d,n}) = \mbox{Discrete}(c_{d,n}|\nu_{d,n})$\\
\hline
\end{tabular}
\end{center}
\end{table*}

\subsubsection{Greedy subtree selection} As mentioned, we perform a greedy algorithm with respect to the variational objective function to determine a subtree from $\T$ for each document. We first describe the algorithm followed by a mathematical representation. Starting from the root node, we sequentially add nodes from $\T$, selecting from those currently ``activated.'' An activated node is one whose parent is contained within the subtree but which is not itself in the subtree. 

To determine which node to add, we look at which node will give the greatest increase in the variational objective when the $q$ distributions for the document-specific beta distributions are fixed to their priors and the variational distribution for each word's topic indicator $q$ distribution ($\nu_{d,n}$ in Table \ref{tab.q_dist}) is zero on the remaining unactivated nodes. That is, we then ask the question: Which of the activated nodes not currently in the subtree will lead to the greatest increase in the variational objective under this restricted $q$ distribution? 

The reason we consider this restricted distribution is that there is a closed form calculation for each node, and so no iterations are required in this step and the algorithm is much faster. Calculating this score only involves optimizing the variational parameter $\nu_{d,n}$ for each word over the current subtree plus the candidate node. We continue adding the maximizing node until the marginal increase in the objective falls below a threshold. We give a more formal description of this below.

\paragraph{\it Coordinate update for $q(z^{(d)}_{\ii,j})$} As defined in Table \ref{tab.q_dist}, $z^{(d)}_{\ii,j}$ is the variable that indicates the index of the atom from the global DP $G_{\ii}$ pointed to by the $j$th stick-breaking weight in $G^{(d)}_{\ii}$. We select a delta $q$ distribution for this variable, meaning we make a hard assignment for this value. These values also define the subtree for document $d$. Starting with an empty tree, all atoms in $G_{\ii_0}$ constitute the activated set. Adding the first node is equivalent to determining the value for $z^{(d)}_{\ii_0,1}$; in general, creating a subtree for $\T_d$, which we denote as $\T'_d$, is equivalent to determining which $z^{(d)}_{\ii,j}$ to include in $\T'_d$ and the atoms to which they point.

For a subtree of size $t$ corresponding to document $d$, let the set $\mathcal{I}_{d,t}$ contain the index values of the included nodes, let $\mathcal{S}_{d,t} = \{\ii\, :\, pa(\ii)\in\mathcal{I}_{d,t}, \ii \not\in \mathcal{I}_{d,t}\}$ be the set of candidate nodes to add to $\mathcal{T}'$. Then provided the marginal increase in the variational objective is above a preset threshold, we increment the subtree by letting $\mathcal{I}_{d,t+1} \leftarrow \mathcal{I}_{d,t} \cup \ii^*$, where
$$\ii^* = \arg\max_{\ii'\in \mathcal{S}_{d,t}}\, \sum_{n=1}^{N_d}\, \max_{\nu_{d,n}:\, \mathcal{C}_{d,t,\ii'}}\, \E_q[\ln p(W_{d,n}|c_{d,n},\theta)]\quad\quad\quad$$
\begin{equation}\label{eq.greedy_select}
\quad\quad\quad +~ \E_q[\ln p(c_{d,n},\boldsymbol{z}^{(d)}|V,V_d,U_d)] - \E_q[\ln q(c_{d,n})].
\end{equation}
We let $\mathcal{C}_{d,t,\ii'}$ denote the discussed conditions, that $\nu_{d,n}(\ii)=0$ for all $\ii \not\in \mathcal{I}_{d,t}\cup \ii'$ and that $q(\cdot)$ is fixed to the prior for all other distributions. The optimal values for $\nu_{d,n}$ are given below in Eq.\ (\ref{eqn.upC}). 

We note two aspects of this greedy algorithm. First, though the stick-breaking construction of the document-level DP given in Eq.\ (\ref{eqn.nHDP_stick_second}) allows for atoms to repeat, in this algorithm each additional atom is new, since there is no advantage in duplicating atoms. Therefore, the algorithm approximates each $G^{(d)}_{\ii}$ by selecting and reordering a subset of atoms from $G_{\ii}$ for its stick-breaking construction. (The subtree $\T'_d$ may also contain zero atoms or one atom from a $G_{\ii}$.) The second aspect we point out is the changing prior on the same node in $\T$. If the atom $\theta_{(\ii,m)}$ is a candidate for addition, then it remains a candidate until it is either selected by a $z^{(d)}_{\ii,j}$, or the algorithm terminates. The prior on selecting this atom changes, however, depending on whether it is a candidate for $z^{(d)}_{\ii,j}$ or $z^{(d)}_{\ii,j'}$. Therefore, incorporating a sibling of $\theta_{(\ii,m)}$ 
impacts the prior on incorporating $\theta_{(\ii,m)}$.

\subsubsection{Coordinate updates for document variables}
Given the subtree $\T'_d$ selected for document $d$, we optimize the variational parameters for the $q$ distributions on $c_{d,n}$, $V^{(d)}_{\ii,j}$ and $U_{d,\ii}$ over that subtree.

\paragraph{\it Coordinate update for $q(c_{d,n})$}
The variational distribution on the path for word $W_{d,n}$ is
\begin{equation}\label{eqn.upC}
 \nu_{d,n}(\ii) \propto \exp\left\lbrace\E_q[\ln \theta_{\ii,W_{d,n}}] + \E_q[\ln\pi_{d,\ii}]\right\rbrace,
\end{equation}
where the prior term $\pi_{d,\ii}$ is the tree-structured prior of the nHDP,
$$\pi_{d,\ii} = \left[\prod_{(\ii',i)\subseteq\ii}\textstyle\prod_j\left(V^{(d)}_{\ii',j}\textstyle\prod_{m<j}(1-V^{(d)}_{\ii',m})\right)^{\I(z^{(d)}_{\ii',j}=i)}\right]\quad\quad$$
\begin{equation}
\quad\quad\quad\quad \times ~\left[U_{d,\ii}\prod_{\ii'\subset\ii}(1-U_{d,\ii'})\right].
\end{equation}
We use the notation $\ii'\subset\ii$ to indicate the subsequences of $\ii$ starting from the first value. The expectation $\E_q[\ln \theta_{\ii,w}] = \psi(\lambda_{\ii,w}) - \psi(\sum_w \lambda_{\ii,w})$, where $\psi(\cdot)$ is the digamma function. Also, for a general random variable $Y \sim \Beta(a,b)$, $\E[\ln Y] = \psi(a) - \psi(a+b)$ and $\E[\ln(1-Y)] = \psi(b) - \psi(a+b)$. The corresponding values of $a$ and $b$ for $U$ and $V$ are given in their respective updates below.

We note that this has a familiar feel as LDA, but where LDA uses a flat Dirichlet prior on $\pi_d$, the nHDP uses a prior that is a tree-structured product of beta random variables. Though the form of the prior is more complicated, the independence results in simple closed-form updates for these beta variables that only depend on $\nu_{d,n}$.

\paragraph{\it Coordinate update for $q(V^{(d)}_{\ii,j})$}
The variational parameter updates for the document-level stick-breaking proportions are
\begin{eqnarray}\label{eqn.upUV}
 u^{(d)}_{\ii,j}\hspace{-2mm} &=&\hspace{-2mm} 1 + \sum_{\ii':(\ii,j)\subseteq\ii'} \sum_{n=1}^{N_d} \nu_{d,n}(\ii'),\\
 v^{(d)}_{\ii,j}\hspace{-2mm} &=&\hspace{-2mm} \beta + \sum_{\ii':\ii\subset\ii'}\I\left(\bigcup_{m>j}\{z^{(d)}_{\ii,m}=\ii'(l+1)\}\right)\sum_{n=1}^{N_d} \nu_{d,n}(\ii').\nonumber
\end{eqnarray}
In words, the statistic for the first parameter is the expected number of words in document $d$ that pass through or stop at node $(\ii,j)$. The statistic for the second parameter is the expected number of words from document $d$ whose paths pass through the same parent $\ii$, but then transition to a node with index greater than $j$ according to the indicators $z^{(d)}_{\ii,m}$ from the document-level stick-breaking construction of $G_{\ii}^{(d)}$.

\paragraph{\it Coordinate update for $q(U_{d,\ii})$} The variational parameter updates for the switching probabilities are similar to those of the document-level stick-breaking process, but collect the statistics from $\nu_{d,n}$ in a slightly different way,
\begin{eqnarray}
 a_{d,\ii} &=& \gamma_1 + \sum_{n=1}^{N_d} \nu_{d,n}(\ii),\\\label{eqn.upAB}s
 b_{d,\ii} &=& \gamma_2 + \sum_{\ii':\ii\subset\ii'}\sum_{n=1}^{N_d}\nu_{d,n}(\ii').
\end{eqnarray}
In words, the statistic for the first parameter is the expected number of words that use the topic at node $\ii$. The statistic for the second parameter is the expected number of words that pass through node $\ii$ but do not terminate there.

\subsubsection{Stochastic updates for corpus variables}
After selecting the subtrees and updating the local document-specific variational parameters for each document $d$ in sub-batch $s$, we take a step in the direction of the natural gradient of the parameters of the $q$ distributions on the global variables. These include the topics $\theta_{\ii}$ and the global stick-breaking proportions $V_{\ii_l,j}$.

\paragraph{\it Stochastic update for $q(\theta_{\ii})$} For the stochastic update of the Dirichlet $q$ distributions on each topic $\theta_{\ii}$, first form the vector $\lambda'_{\ii}$ of sufficient statistics using the data in sub-batch $s$,
\begin{equation}\label{eqn.upTheta}
\lambda'_{\ii,w} = \frac{D}{|C_s|}\sum_{d\in C_s}\sum_{n=1}^{N_d} \nu_{d,n}(\ii)\I\{W_{d,n} = w\},
\end{equation}
for $w=1,\dots,\mathcal{V}.$ This vector contains the expected number of words with index $w$ that originate from topic $\theta_{\ii}$ over documents indexed by $C_s$. According to the discussion on stochastic inference in Section \ref{sec.stoch}, we scale this to a corpus of size $D$. The update for the associated $q$ distribution is
\begin{equation}
 \lambda^{s+1}_{\ii,w} = \lambda_0 + (1-\rho_s)\lambda^s_{\ii,w} + \rho_s \lambda'_{\ii,w}.
\end{equation}
We see a blending of the old statistics with the new in this update. Since $\rho_s \rightarrow 0$ as $s$ increases, the algorithm uses less and less information from new sub-groups of documents, which reflects the increasing confidence in this parameter value as more data is seen.

\paragraph{\it Stochastic update for $q(V_{\ii_l,j})$} As with $\theta_{\ii}$, we first collect the sufficient statistics for the $q$ distribution on $V_{\ii_l,j}$ from the documents in sub-batch $s$,
\begin{eqnarray}
\tau'_{\ii_l,j} &=& \frac{D}{|C_s|}\sum_{d\in C_s} \I\{\ii_l\in\mathcal{I}_d\},\\
\tau''_{\ii_l,j} &=& \frac{D}{|C_s|}\sum_{d\in C_s} \sum_{j>i_l}  \I\{(pa(\ii_l),j)\in\mathcal{I}_d\}.
\end{eqnarray}
The first value scales up the number of documents in sub-batch $s$ that include atom $\theta_{(\ii,j)}$ in their subtree; the second value scales up the number of times an atom of higher index value in the same DP is used by a document in sub-batch $s$. The update to the global variational parameters are
\begin{eqnarray}
 \tau_{\ii_l,j}^{(1)}(s+1) &=& 1 + (1-\rho_s)\tau_{\ii_l,j}^{(1)}(s) + \rho_s \tau'_{\ii_l,j},\\\label{eqn.upTau}
 \tau_{\ii_l,j}^{(2)}(s+1) &=& \alpha + (1-\rho_s)\tau_{\ii_l,j}^{(2)}(s) + \rho_s \tau''_{\ii_l,j}.
\end{eqnarray}
Again, we see a blending of old information with new.

\begin{table*}
\begin{center}\caption{Comparison of the nHDP with the nCRP in the batch inference setting using the predictive log likelihood.}\label{tab.q_dist}
\begin{tabular}{|c|ccc|}
\hline
Method$\backslash$Dataset & JACM & Psych. Review & PNAS\\
\hline
Variational nHDP & -5.405 $\pm$ 0.012 & -5.674 $\pm$ 0.019 & -6.304 $\pm$ 0.003\\
Variational nCRP (Wang, et al.\ \cite{Wang:2009}) & -5.433 $\pm$ 0.010 & -5.843 $\pm$ 0.015 & -6.574 $\pm$ 0.005\\
Gibbs nCRP (Wang, et al.\ \cite{Wang:2009}) 	 & -5.392 $\pm$ 0.005 & -5.783 $\pm$ 0.015 & -6.496 $\pm$ 0.007\\
\hline
\end{tabular}
\end{center}\vspace{-10pt}
\end{table*}
\section{Experiments}\label{sec.experiments}
We present an empirical evaluation of the nested HDP topic model in the stochastic and the batch inference settings. We first present batch results on three smaller data sets to verify that our multi-path approach gives an improvement over the single-path nested CRP. We then move to the stochastic inference setting, where we perform experiments on 1.8 million documents from \emph{The New York Times} and 2.7 million documents from \emph{Wikipedia}. We compare with other recent stochastic inference algorithms for topic models: stochastic LDA \cite{Hoffman:2010} and the stochastic HDP \cite{Wang:2011}. As is fairly standard with the optimization-based variational inference, we use truncated stick-breaking processes for all DPs \cite{Blei:2005}\cite{Kurihara:2006}. With this method, we truncate the {\it posterior} approximation by not allowing words to come from topics beyond the truncation index (i.e., fixing $c_{d,n}((\ii,j)) = 0$ for all $j > n$). The truncation is set to something reasonably large, and the 
posterior inference procedure then shrinks the number of used topics to something smaller than the number provided. In our large-scale experiments, we truncate to $n_1 = 20$ first level nodes, $n_2 = 10$ children for each of these nodes and $n_3 = 5$ children of each of these second level nodes. We consider three level trees, corresponding intuitively to ``general'', ``specific'' and ``specialized'' levels of words. Though the nHDP is nonparametric in level as well, we are more interested in the nonparametric aspect of the Dirichlet process here.

\subsection{Initialization}\label{sec.init}
Before presenting our results, we discuss our method for initializing the topic distributions of the tree. As with most Bayesian models, inference for hierarchical topic models can benefit greatly from a good initialization. Our goal is to find a method for quickly centering the posterior mean of each topic so that they contain some information about their hierarchical relationships. We briefly discuss our approach for initializing the global variational topic parameters $\lambda_{\ii}$ of the nHDP.

Using a small set of documents (e.g, 10,000) from the training set, we form the empirical distribution for each document on the vocabulary. We then perform k-means clustering of these probability vectors using the $L_1$ distance measure (i.e., total variation). At the top level, we partition the data into $n_1$ groups, corresponding to $n_1$ children of the root node from the truncated stick-breaking process. We then subtract the mean of a group (a probability vector) from all data within that group, set any negative values to zero and renormalize. We loosely think of this as the ``probability of what remains''---a distribution on words not captured by the parent distributions. Within each group we again perform k-means clustering, obtaining $n_2$ probability vectors for each of the $n_1$ groups, and again subtracting, setting negative values to zero and renormalizing the remainder of each probability vector for a document.

Through this hierarchical k-means clustering, we obtain $n_1$ probability vectors at the top level, $n_2$ probability vectors beneath each top-level vector for the second level, $n_3$ probability vectors beneath each of these second-level vectors, etc. The $n_i$ vectors obtained from any sub-group of data are refinements of an already coherent sub-group of data, since that sub-group is itself a cluster from a larger group. Therefore, the resulting tree will have some thematic coherence. The clusters from this algorithm are used to initialize the nodes within the nHDP tree. For a mean probability vector $\hat{\lambda}_{\ii}$ obtained from this algorithm, we set the corresponding variational parameter for the topic Dirichlet distribution $q$ to $\lambda_{\ii} = N(\kappa\hat{\lambda}_{\ii} + (1-\kappa)(\boldsymbol{1}/\mathcal{V}+v_{\ii}))$ for $\kappa \in [0,1]$, $N$ a scaling factor and $v_{\ii} \iid \mbox{Dirichlet}(100\boldsymbol{1}_{\mathcal{V}}/\mathcal{V})$. This initializes the mean of $\theta_{\ii}$ to 
be slightly peaked around $\hat{\lambda}_{\ii}$, while the uniform vector and $\kappa$ help determine  the variance and $v_{\ii}$ provides some randomness. In our algorithms we set $\kappa = 0.5$ and $N$ equal to the number of documents.

\subsection{A batch comparison}
Before comparing our stochastic inference algorithm for the nHDP with similar algorithms for LDA and the HDP, we compare a batch version with the nCRP on three smaller data sets. This will verify the advantage of giving each document access to the entire tree versus forcing each document to follow one path. We compare the variational nHDP topic model with both the variational nCRP \cite{Wang:2009} and the Gibbs sampling nCRP \cite{Blei:2010}, using the parameter settings in those papers to facilitate comparison. We consider three corpora for our experiments: ($i$) The \emph{Journal of the ACM}, a collection of 536 abstracts from the years 1987--2004 with vocabulary size 1,539; ($ii$) The \emph{Psychological Review}, a collection of 1,272 abstracts from the years 1967--2003 with vocabulary size 1,971; and ($iii$) The \emph{Proceedings of the National Academy of Science}, a collection of 5,000 abstracts from the years 1991--2001 with a vocabulary size of 7,762. The average number of words per document for the 
three corpora are 45, 108 and 179, respectively.

As mentioned, variational inference for Dirichlet process priors uses a truncation of the variational distribution, which limits the number of topics that are learned \cite{Blei:2005}\cite{Kurihara:2006}. This truncation is set to a number larger than the anticipated number of topics necessary for modeling the data set, but can be increased if more are needed \cite{Wang:2012}. We use a truncated tree of $(10,7,5)$ for modeling these corpora, where 10 children of the root node each have 7 children, which themselves each have 5 children for a total of 420 nodes. Because these three data sets contain stop words, we follow \cite{Wang:2009} and \cite{Blei:2010} by including a root node shared by all documents for this batch problem only. Following \cite{Wang:2009}, we perform five-fold cross validation to evaluate performance on each corpus.

We present our results in Table \ref{tab.q_dist}, where we show the predictive log likelihood on a held-out test set. We see that for all data sets, the variational nHDP outperforms the variational nCRP. For the two larger data sets, the variational nHDP also outperforms Gibbs sampling for the nCRP. Given the relative sizes of these corpora, we see that the benefit of learning a per-document distribution on the full tree rather than a shared distribution on paths appears to increase as the corpus size and document size increase. Since we are interested in the ``Big Data'' regime, this strongly hints at an advantage of our nHDP approach over the nCRP. We omit a comparison with Gibbs nHDP since MCMC methods are not amenable to large data sets for this problem.

\begin{figure}[t]
 \includegraphics[width=.45\textwidth]{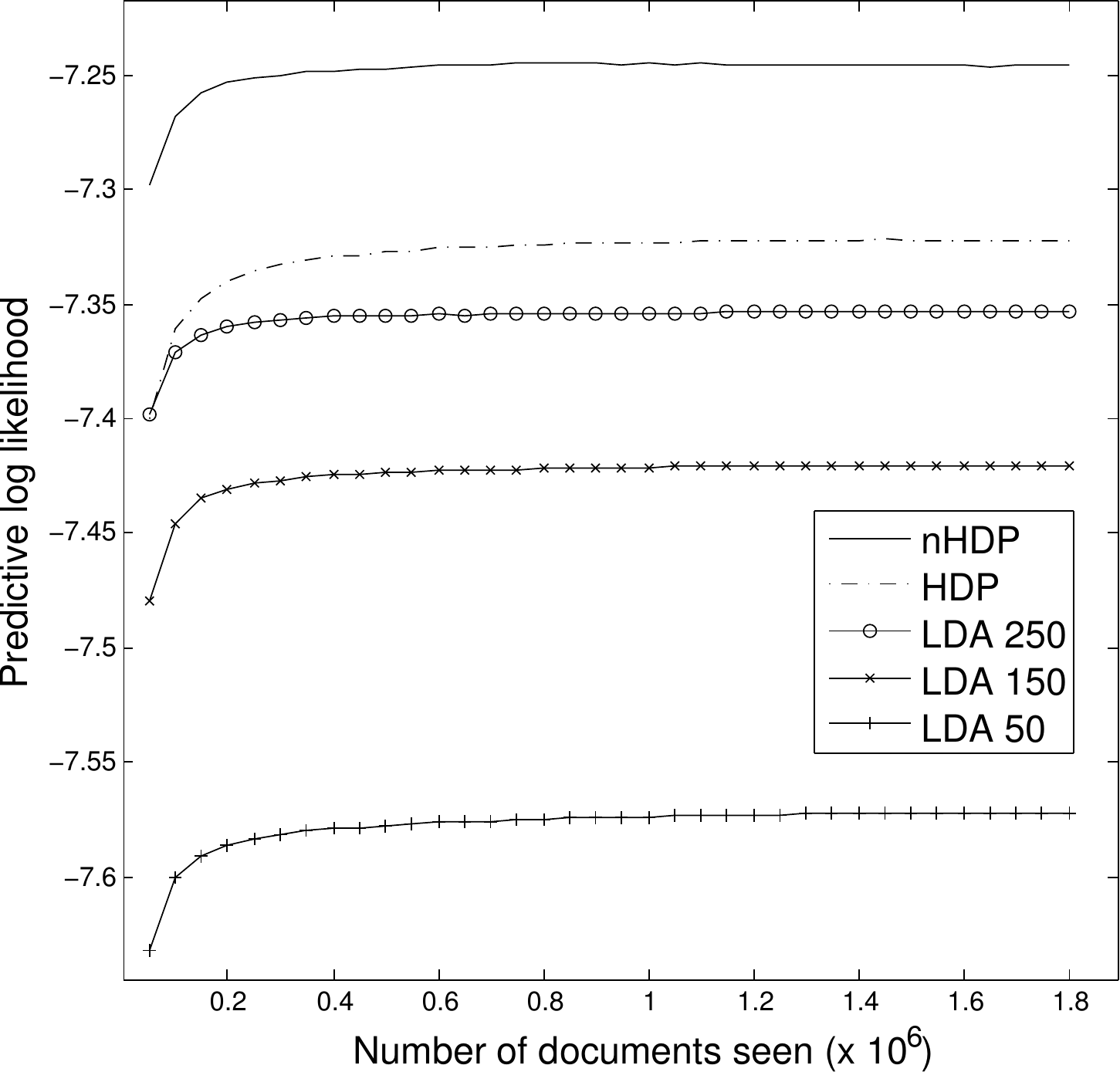}
\caption{The New York Times: Average predictive log likelihood on a held-out test set as a function of training documents seen.\label{fig.nyt_llik}}
\end{figure}
\begin{figure}
 \includegraphics[width=.45\textwidth]{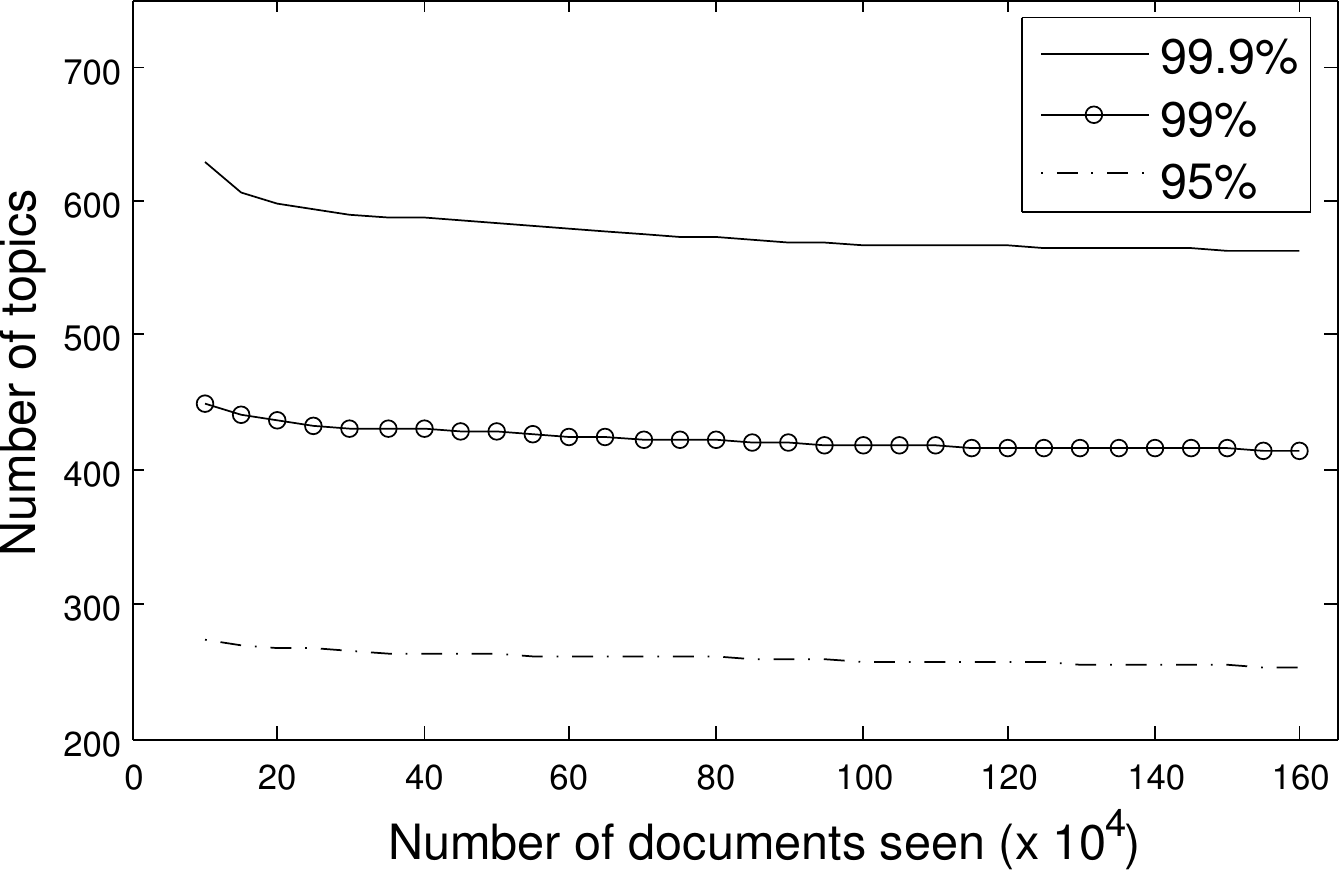}
\caption{New York Times: The total size of the tree as a function of documents seen. We show the smallest number of nodes containing 95\%, 99\% and 99.9\% of the posterior mass.\label{fig.nyt_topic_count}}
\vspace{10pt}
\includegraphics[width=.485\textwidth]{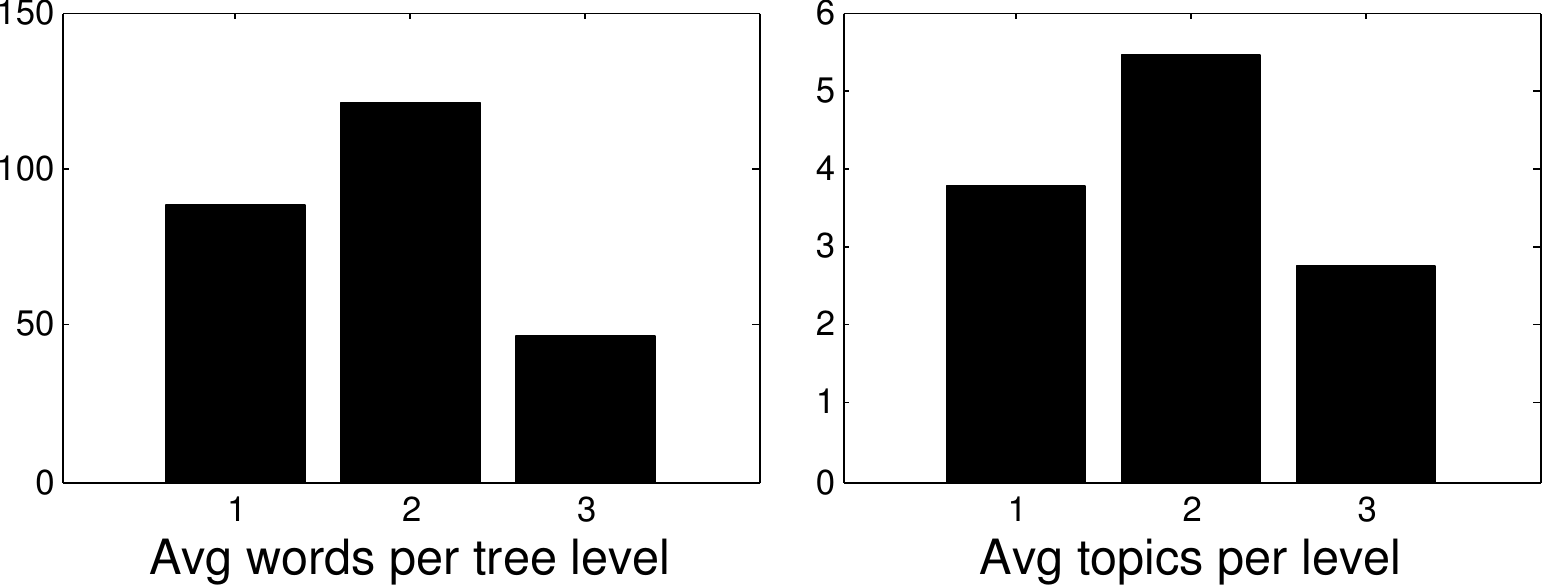}
\caption{The New York Times: Per-document statistics from the test set using the tree at the final step of the algorithm. (left) The average number of words per tree level. (right) The average number of nodes per level with more than one expected observation.\label{fig.nyt_breakdown}}
\vspace{10pt}
 \includegraphics[width=.45\textwidth]{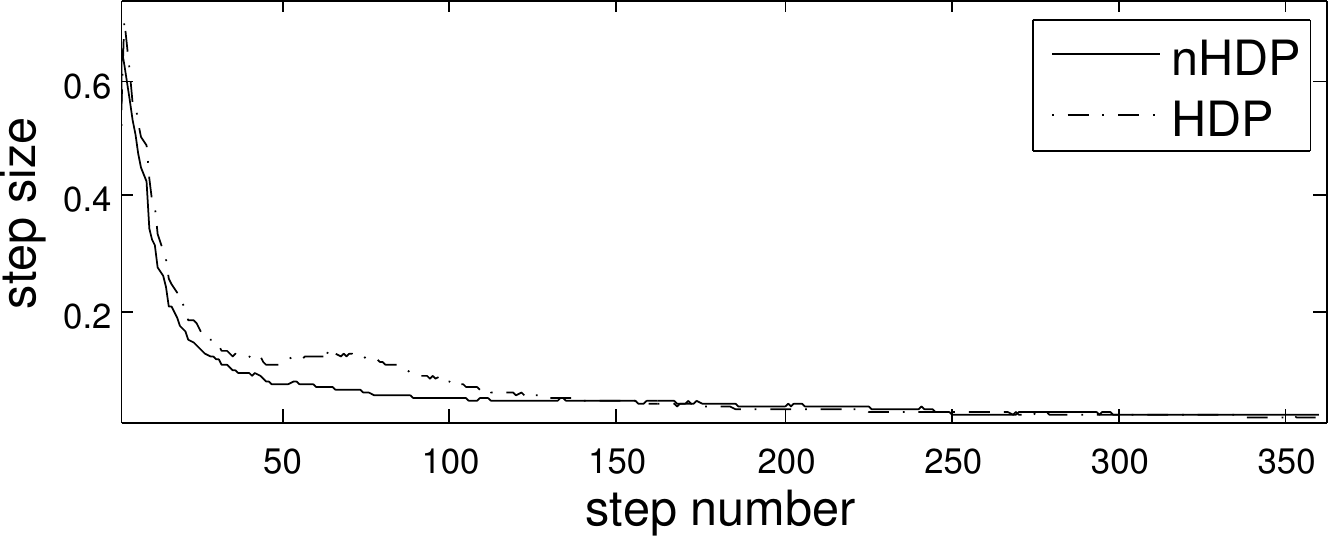}
\caption{New York Times: The adaptively learned step size.\label{fig.nyt_rho}}
\end{figure}\vspace{-10pt}

\subsection{Stochastic inference for large corpora}

We next present an evaluation of our stochastic variational inference algorithm on \emph{The New York Times} and \emph{Wikipedia}. These are both very large data sets, with \emph{The New York Times} containing roughly 1.8 million articles and \emph{Wikipedia} roughly 2.7 million web pages. The average document size is somewhat larger than those considered in our batch experiments as well, with an article from \emph{The New York Times} containing 254 words on average taken from a vocabulary size of 8,000, and \emph{Wikipedia} 164 words on average taken from a vocabulary size of 7,702. For this problem we remove stop words and rare words.

\begin{figure*}[t!]\centering
 \includegraphics[width=.95\textwidth]{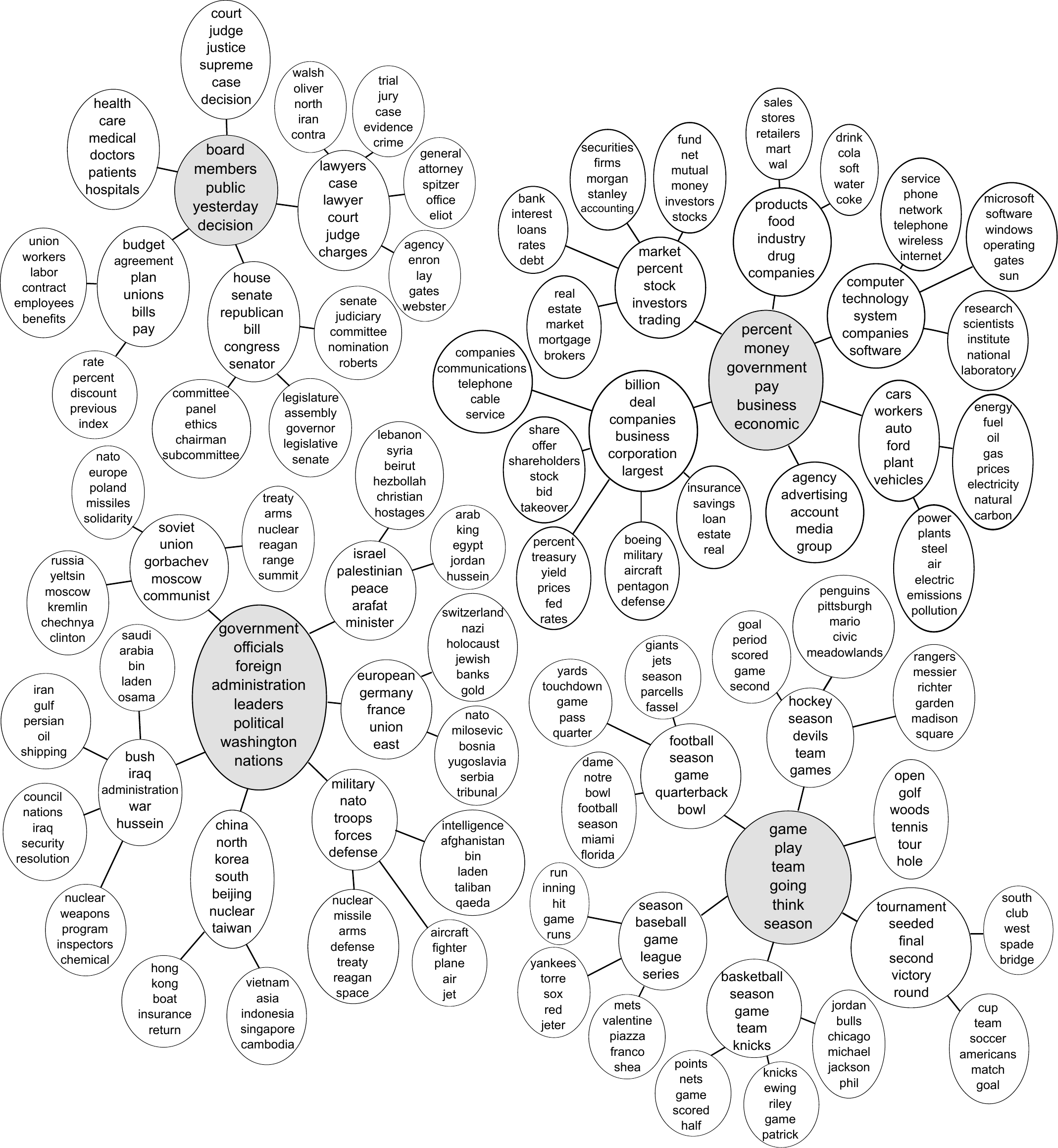}
\caption{Tree-structured topics from The New York Times. The shaded node is the top-level node and lines indicate dependencies within the tree. In general, topics are learning in increasing levels of specificity. For clarity, we have removed grammatical variations of the same word, such as ``scientist'' and ``scientists.''\label{fig.nyt_topics}}
\end{figure*}

\subsubsection{Setup} We use the algorithm discussed in Section \ref{sec.init} to initialize a three-level tree with $(20,10,5)$ child nodes per level, giving a total of 1,220 initial topics. For the Dirichlet processes, we set all top-level DP concentration parameters to $\alpha = 5$ and the second-level DP concentration parameters to $\beta = 1$. For the switching probabilities $U$, we set the beta distribution hyperparameters for the tree level prior to $\gamma_1 = 1/3$ and $\gamma_2 = 2/3$, slightly encouraging a word to continue down the tree. We set the base Dirichlet parameter $\lambda_0 = 0.1$. For our greedy subtree selection algorithm, we stop adding nodes to the subtree when the marginal improvement to the lower bound falls below $10^{-3}$. When optimizing the local variational parameters of a document given its subtree, we continue iterating until the fractional change in the $L_1$ distance of the empirical distribution of words falls below $10^{-2}$.

We hold out a data set for each corpus for testing, 14,268 documents for testing \emph{The New York Times} and 8,704 documents for testing \emph{Wikipedia}. To quantitatively assess the performance, at various points in the learning process we calculate the predictive log likelihood on a fraction of the test set as follows: Holding the top-level variational parameters fixed, for each test document we randomly partition the words into a 90/10 percent split. We then learn document-specific variational parameters for the 90\% portion. Following \cite{Teh:2008}\cite{Wang:2009}, we use the mean of each $q$ distribution to form a predictive distribution for the remaining words of that document. With this distribution, we calculate the average predictive log likelihood of the 10\% portion to assess performance. For comparison, we evaluate stochastic inference algorithms for LDA and the HDP in the same manner. In all algorithms, we use an algorithm for adaptively learning the step size $\rho_s$ as presented by 
Ranganath, et al.\ \cite{Ranganath:2013}.

\begin{figure}
 \includegraphics[width=.45\textwidth]{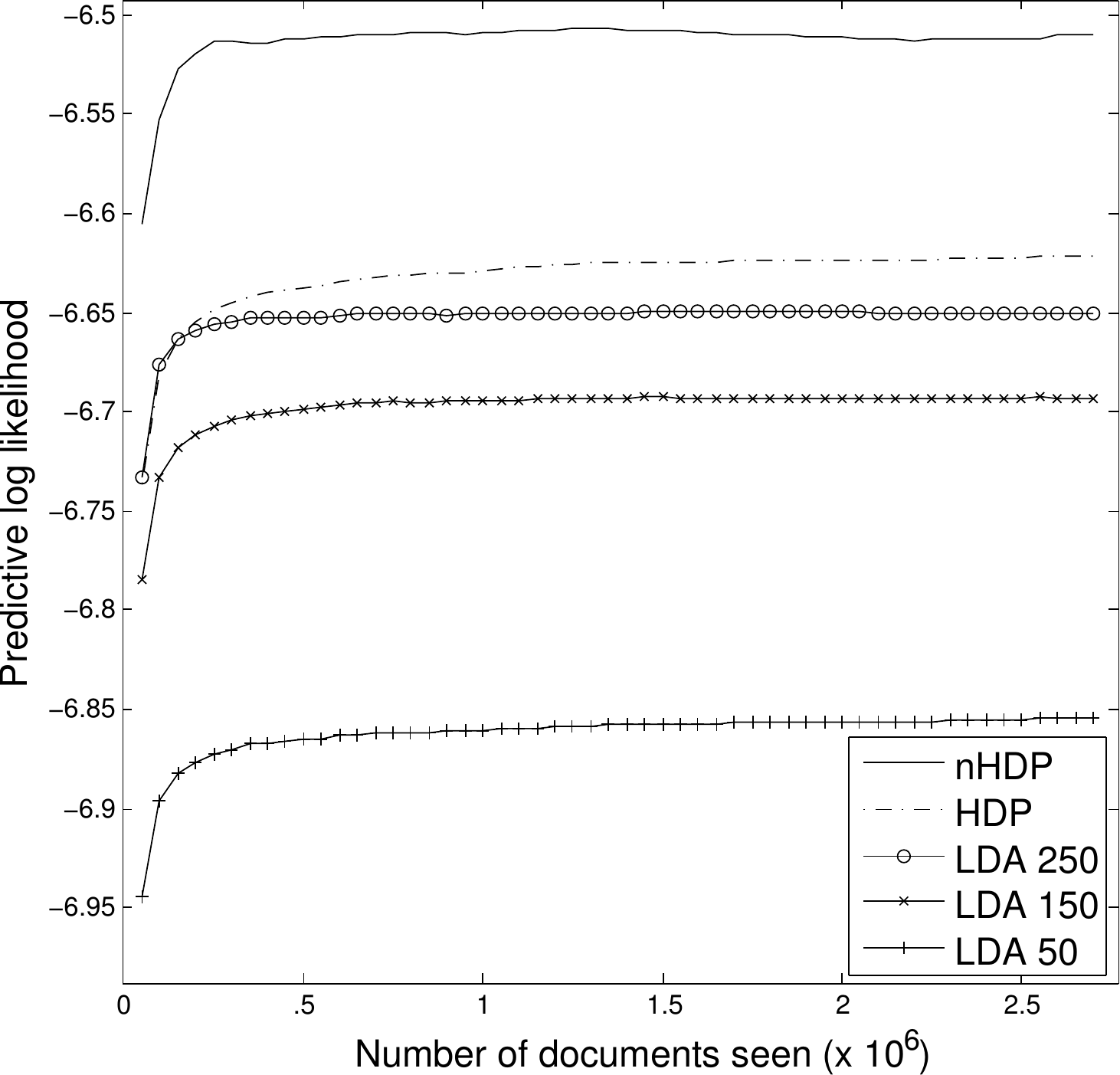}
\caption{Wikipedia: Average predictive log likelihood on a held-out test set as a function of training documents seen.\label{fig.wiki_llik}}
\end{figure}
\begin{figure}
 \includegraphics[width=.45\textwidth]{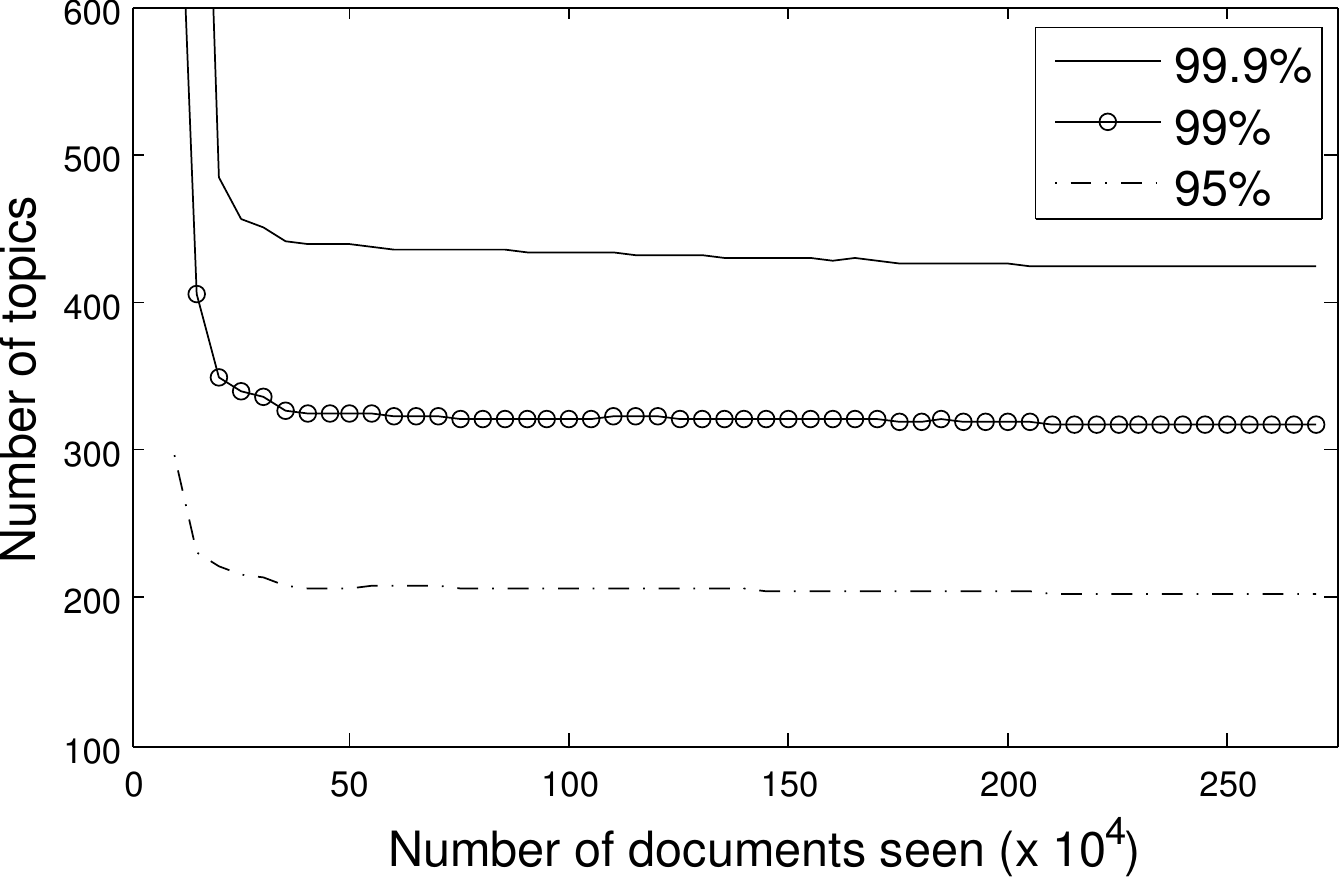}
\caption{Wikipedia: The total size of the tree as a function of documents seen. We show the smallest number of nodes containing 95\%, 99\% and 99.9\% of the posterior mass.\label{fig.wiki_topic_count}}
\vspace{10pt}
 \includegraphics[width=.485\textwidth]{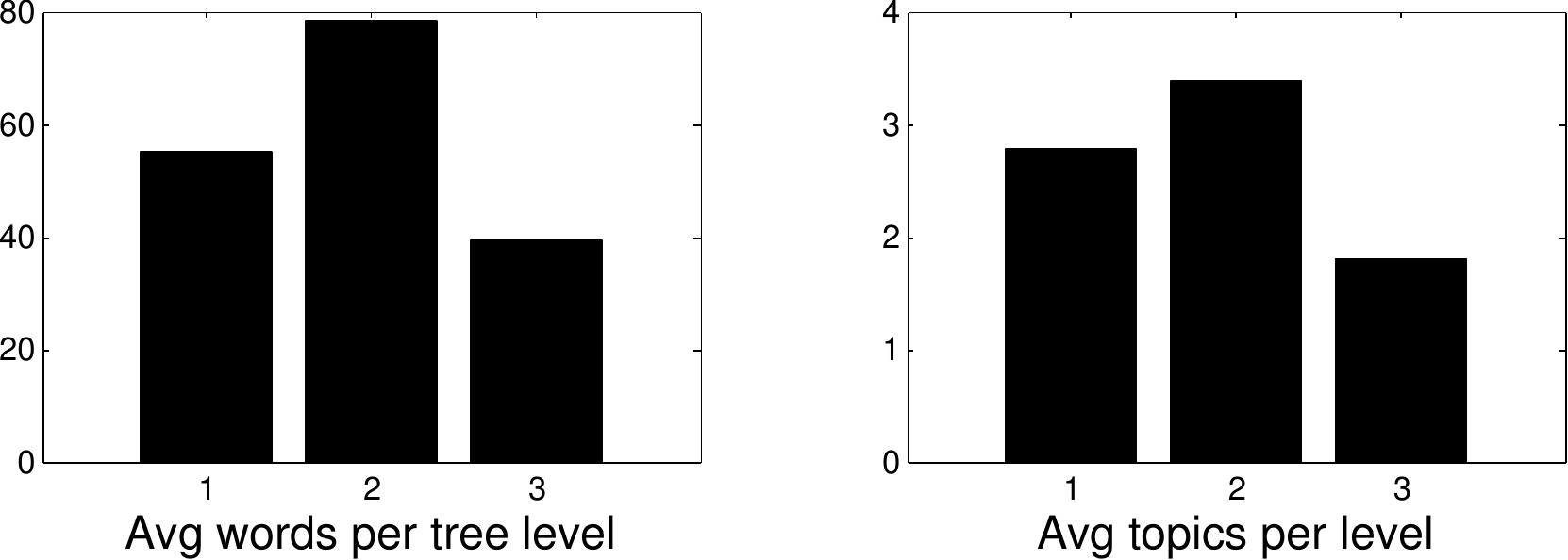}
\caption{Wikipedia: Per-document statistics from the test set using the tree at the final step of the algorithm. (left) The average number of words per tree level. (right) The average number of nodes per level with more than one expected observation.\label{fig.wiki_breakdown}}
\vspace{10pt}
 \includegraphics[width=.45\textwidth]{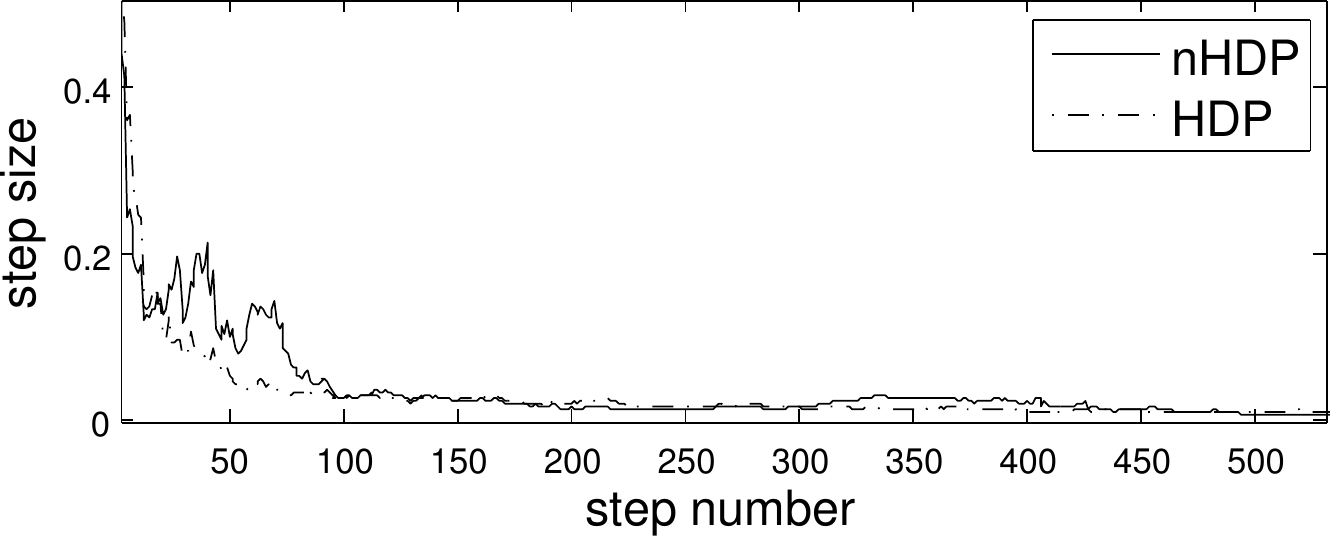}
\caption{Wikipedia: The adaptively learned step size.\label{fig.wiki_rho}}
\end{figure}
\begin{figure}\centering
 \includegraphics[width=.5\textwidth]{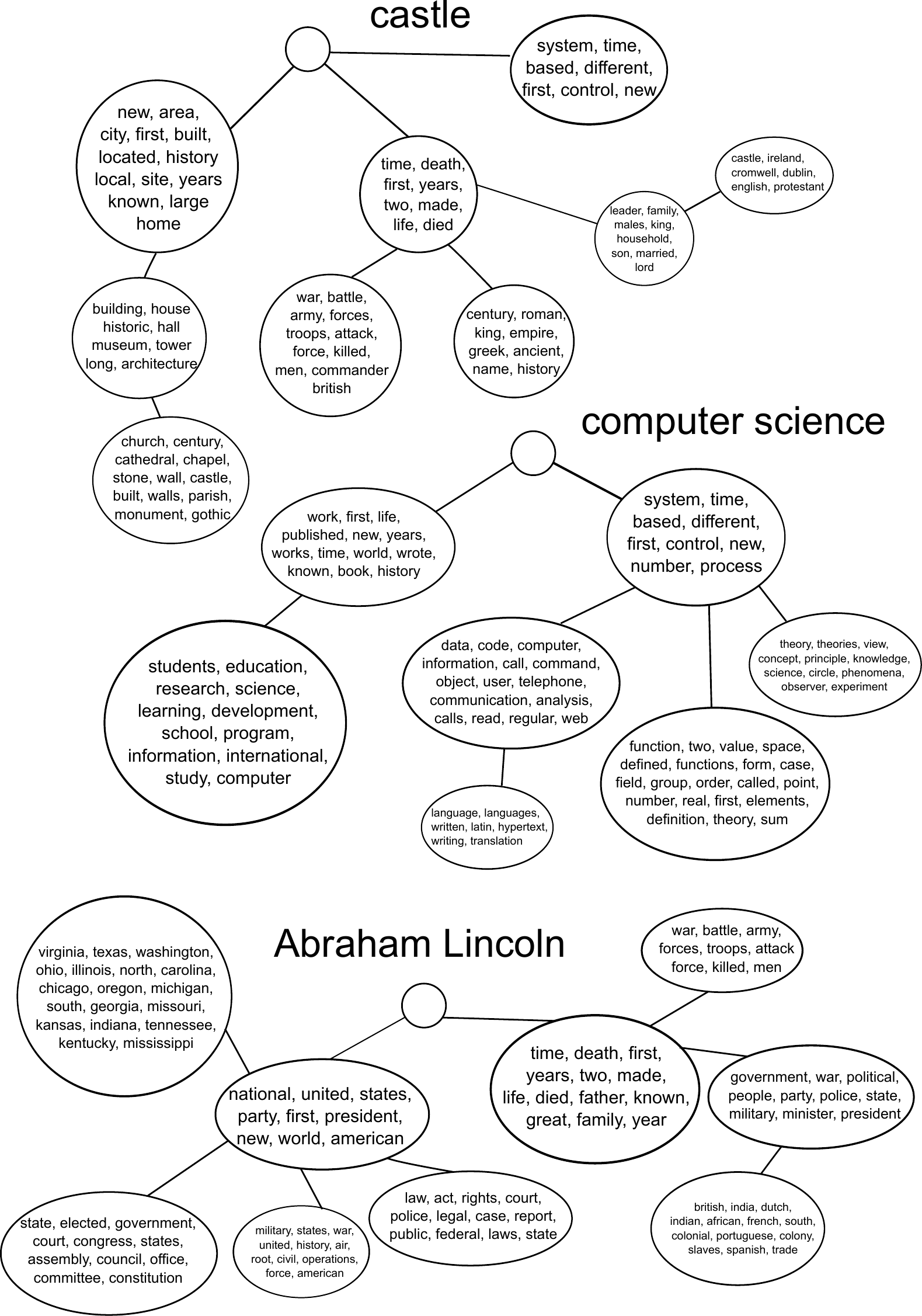}
\caption{Examples of subtrees for three articles from \emph{Wikipedia}. The three sizes of font indicate differentiate the more probable topics from the less probable.\label{fig.topic_paths}}
\end{figure}

\subsubsection{The New York Times}
We first present our results for \emph{The New York Times}. In Figure \ref{fig.nyt_llik} we show the average predictive log likelihood on unseen words as a function of the number of documents processed during model learning. We see an improvement in performance as the amount of data processed increases. We also note an improvement in the performance of the nHDP compared with LDA and the HDP. In Figure \ref{fig.nyt_topic_count}  we give a sense of the size of the tree as a function of documents seen. Since all topics aren't used equally, we show the minimum number of nodes containing $95\%$, $99\%$ and $99.9\%$ of all data in the posterior.  In Figure \ref{fig.nyt_breakdown} we show document-level statistics from the test set at the final step of the algorithm. These include the word allocations by level and the number of topics used per level. We note that while the tree has three levels, roughly 12 topics are being used (in varying degrees) per document. This is in contrast to the three topics that would be 
available to any document with the nCRP. Thus there is a clear advantage in allowing each document to have access to the entire tree. We show the adaptively learned step size in Figure \ref{fig.nyt_rho}.

In Figure \ref{fig.nyt_topics} we show example topics from the model and their relative structure. For each node we show the most probable words according to the approximate posterior $q$ distribution of the topic. We show four topics from the top level of the tree (shaded), and connect topics according to parent/child relationship. The model learns a meaningful hierarchical structure; for example, the sports subtree branches into the various sports, which themselves appear to branch by teams. In the foreign affairs subtree, children tend to group by major subregion and then branch out into subregion or issue. If a sports document incorporated topics on foreign affairs, the nHDP would allow words to split into both parts of the tree, but with the nCRP a document would have to pick one or the other, and so a tree could not be learned that distinguished topics with this level of precision.

The algorithm took roughly 20 hours to make one pass through the data set using a single desktop computer, which was sufficient for the model to converge to a set of topics. Runtime for {\it Wikipedia} was comparable.

\subsubsection{Wikipedia}
We show similar results for \emph{Wikipedia} as for \emph{The New York Times}. In Figures \ref{fig.wiki_llik}, \ref{fig.wiki_topic_count}, \ref{fig.wiki_breakdown} and \ref{fig.wiki_rho} we show results corresponding to Figures \ref{fig.nyt_llik}, \ref{fig.nyt_topic_count}, \ref{fig.nyt_breakdown} and \ref{fig.nyt_rho}, respectively for \emph{The New York Times}. We again see an improvement in performance for the nHDP over LDA and the HDP, as well as the increased usage of the tree with the nHDP than would be available in the nCRP.

In Figure \ref{fig.topic_paths}, we see example subtrees used by three documents. We note that the topics contain many more function words than for \emph{The New York Times}, but an underlying hierarchical structure is uncovered that would be unlikely to arise along one path, as the nCRP would require. As with {\it The New York Times}, we see the nonparametric nature of the model in Figure \ref{fig.wiki_topic_count}. Though the model has an 1,220 initial nodes, a small subset are ultimately used by the data.

\subsubsection{Sensitivity analysis}
We present a brief sensitivity analysis of some parameters of the nHDP topic model using the {\it Wikipedia} corpus. In general, we find that the results were not sensitive to the parameter $\lambda_0$ of the base Dirichlet distribution, which is consistent with \cite{Hoffman:2012}. We note that this is typically not the case for topic models, but because of the massive quantity of data we are working with, the data overwhelms the prior in this case. This was similarly found with the global DP parameter $\alpha$.

\begin{figure}[t!]
 \includegraphics[width=.48\textwidth]{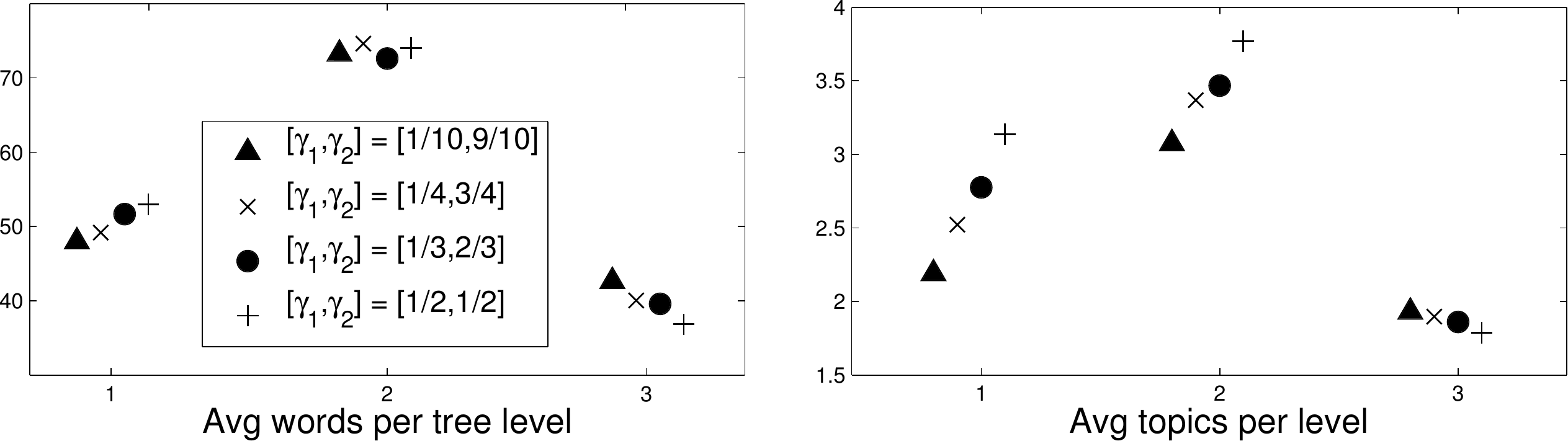}
\caption{Wikipedia: Sensitivity to parameter vector $(\gamma_1,\gamma_2)$ for the stochastic switches. We show the results from Figure \ref{fig.wiki_breakdown} for different settings with $\beta = 1$. \label{fig.sensitivity1}}
\end{figure}

\begin{figure}[t!]
 \includegraphics[width=.48\textwidth]{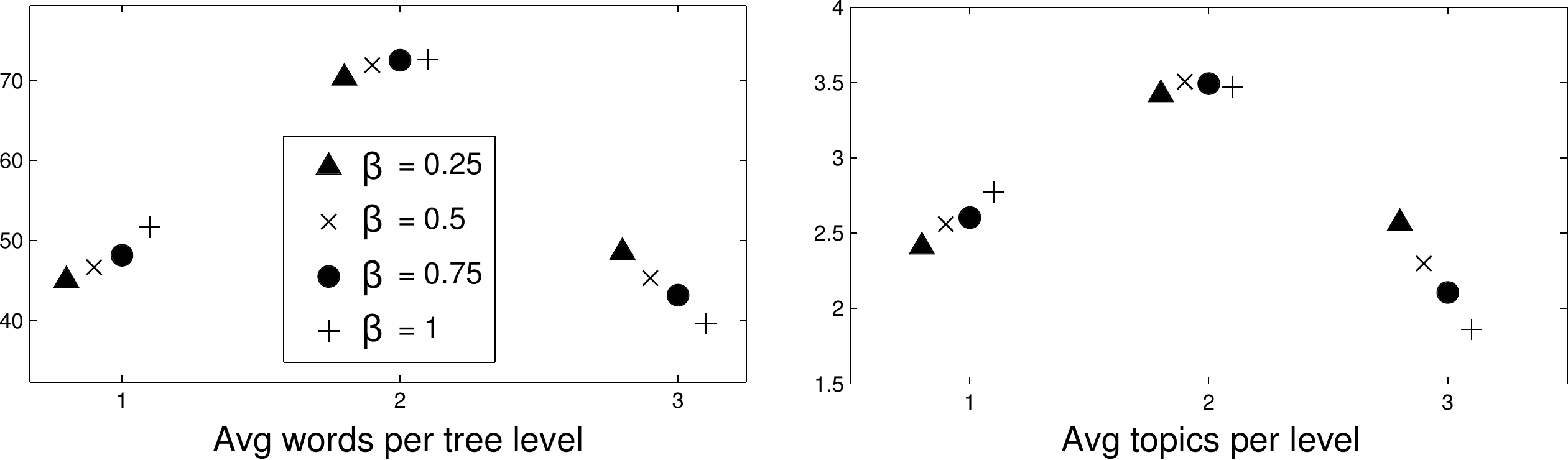}
\caption{Wikipedia: Sensitivity to parameter vector $\beta$ for the local DPs. We show the results from Figure \ref{fig.wiki_breakdown} for different settings and $\gamma_1=1/3, \gamma_2 = 2/3$. \label{fig.sensitivity2}}
\end{figure}

\begin{figure}[t!]
 \includegraphics[width=.48\textwidth]{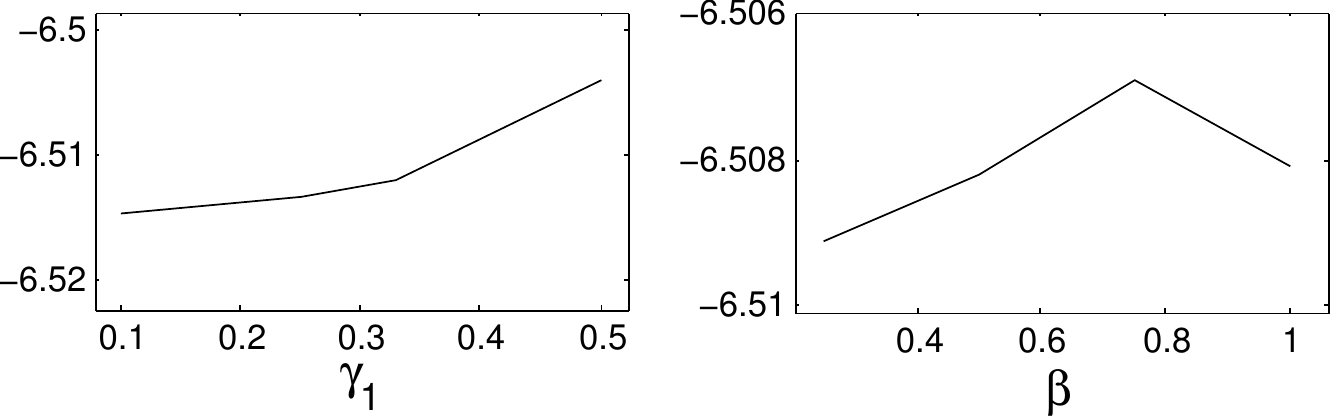}
\caption{Wikipedia: Sensitivity to $(\gamma_1, \gamma_2 = 1-\gamma_2)$ with $\beta=1$ (left), and $\beta$ with $\gamma_1=1/3, \gamma_2=2/3$ (right). We show the predictive log likelihood on the test set. \label{fig.sensitivity3}}
\end{figure}

The document-specific variables have a more significant impact since they only use the data from a single document in their posteriors. In Figures \ref{fig.sensitivity1}--\ref{fig.sensitivity3} we show the sensitivity of the model to the parameters $\beta$ and $(\gamma_1,\gamma_2)$. We consider several values for these parameters, holding $\gamma_1+\gamma_2=1$. As can be seen, the model structure is fairly robust to these values. The tree structure does respond as would be expected from the prior, but there is no major change. The quantitative results in Figure \ref{fig.sensitivity3} indicate that the quality of the model is robust as well. We note that this relative insensitivity is within a parameter range that we believe a priori to be reasonable.


\section{Conclusion}
We have presented the nested hierarchical Dirichlet process (nHDP), an extension of the nested Chinese restaurant process (nCRP) that allows each observation to follow its own path to a topic in the tree. Starting with a stick-breaking construction for the nCRP, the new model samples document-specific path distributions for a shared tree using a nested hierarchy of Dirichlet processes. By giving a document access to the entire tree, we are able to borrow thematic content from various parts of the tree in constructing a document. We developed a stochastic variational inference algorithm that is scalable to very large data sets. We compared the stochastic nHDP topic model with stochastic LDA and HDP and showed how the nHDP can learn meaningful topic hierarchies.

\bibliographystyle{IEEEtran}  
\bibliography{pami_bnp}

\vspace{-30pt}
\begin{IEEEbiography}[{\includegraphics[width=1in,height=1.25in,clip,keepaspectratio]{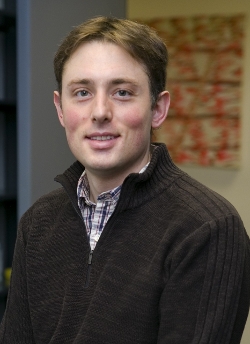}}]{John Paisley}
is an assistant professor in the Department of Electrical Engineering at Columbia University. Prior to that he was a postdoctoral researcher at UC Berkeley and Princeton University. He received the B.S., M.S. and Ph.D. degrees in Electrical Engineering from Duke University in 2004, 2007 and 2010. His research is in the area of machine learning and focuses on developing Bayesian nonparametric models for applications involving text and images.
\end{IEEEbiography}
\vspace{-30pt}
\begin{IEEEbiography}[{\includegraphics[width=1in,height=1.25in,clip,keepaspectratio]{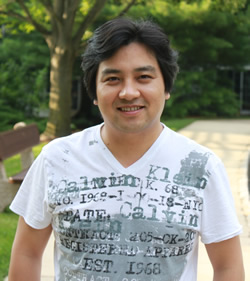}}]{Chong Wang} 
is a Senior Research Scientist in Voleon Capital Management. Before that, he was a project scientist in the Machine Learning department at Carnegie Mellon University. He received his PhD from Princeton University in 2012 in Computer Science.  His thesis was nominated for the ACM Doctoral Dissertation Award by Princeton University. His research focuses on probabilistic graphical models and their applications to real-world problems.
\end{IEEEbiography}
\vspace{-30pt}
\begin{IEEEbiography}[{\includegraphics[width=1in,height=1.25in,clip,keepaspectratio]{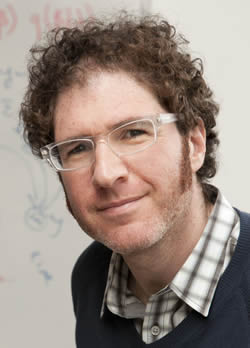}}]{David M. Blei}
is an associate professor of Computer Science at Princeton
University. He received his PhD in 2004 at U.C.\ Berkeley and was a
postdoctoral fellow at Carnegie Mellon University. His research
focuses on probabilistic topic models, Bayesian nonparametric methods,
and approximate posterior inference. He works on a variety of
applications, including text, images, music, social networks, and
scientific data.
\end{IEEEbiography}
\vspace{-30pt}
\begin{IEEEbiography}[{\includegraphics[width=1in,height=1.25in,clip,keepaspectratio]{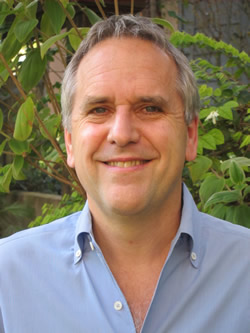}}]{Michael I. Jordan}
 is the Pehong Chen Distinguished Professor in the 
Department of Electrical Engineering and Computer Science and the 
Department of Statistics at the University of California, Berkeley.  
His research in recent years has focused on Bayesian nonparametric 
analysis, probabilistic graphical models, spectral methods, kernel 
machines and applications to problems in statistical genetics, signal 
processing, computational biology, information retrieval and natural 
language processing.  Prof. Jordan is a member of the National 
Academy of Sciences, a member of the National Academy of Engineering
and a member of the American Academy of Arts and Sciences.  He is a 
Fellow of the American Association for the Advancement of Science.  
He has been named a Neyman Lecturer and a Medallion Lecturer by the 
Institute of Mathematical Statistics.  He is an Elected Member of
the International Institute of Statistics.  He is a Fellow of the AAAI,
ACM, ASA, CSS, IMS, IEEE and SIAM.
\end{IEEEbiography}

\end{document}